\documentclass[10pt,twocolumn,letterpaper]{article}

 \usepackage{iccv}              

\usepackage{graphicx}
\usepackage{amsmath}
\usepackage{amssymb}
\usepackage{booktabs}
\usepackage{tabularx}
\usepackage{bbold}
\usepackage{textcomp}  
\usepackage{scalerel}  
\usepackage{caption}
\usepackage{comment}
\usepackage{pifont}
\usepackage{wasysym}
\usepackage{graphicx}

\usepackage{bbding}

\usepackage{tcolorbox}
\usepackage{wrapfig,lipsum}

\usepackage{color}
\usepackage{colortbl}
\usepackage[table]{xcolor}   
\usepackage{xparse, pgf} 

\usepackage{multirow}

\usepackage{xparse}

\usepackage[dvipsnames]{xcolor}

\usepackage{soul}

\definecolor{weakorange}{RGB}{255,230,200} 
\definecolor{weakgray}{RGB}{240,240,240} 
\definecolor{colbest}{rgb}{0.1, 0.6, 0.1}
\definecolor{colworst}{rgb}{0.75, 0, 0}
\definecolor{oursrow}{HTML}{EBF5F8}

\definecolor{yellowhighlight}{HTML}{FFF2CC}
\definecolor{purplehighlight}{HTML}{efdce5}
\definecolor{lightpinkhighlight}{HTML}{FBE5E5}
\definecolor{lightgreenhighlight}{HTML}{E6FAE0}

\NewDocumentCommand{\coloredCell}{m}{%
  \pgfmathparse{(#1>100 ? 100 : (#1<0 ? 0 : #1))}%
  \edef\clamped{\pgfmathresult}%
  \pgfmathparse{\clamped/100.0}%
  \edef\ratio{\pgfmathresult}

  \pgfmathparse{1-\ratio}%
  \edef\gr{\pgfmathresult}%
  \pgfmathparse{1-\ratio}%
  \edef\bl{\pgfmathresult}%

  \edef\myRGB{1,\gr,\bl}%

  \cellcolor[rgb]{\myRGB}%

  \ifdim #1pt>100pt
    100+%
  \else
    #1%
  \fi
}

\newcommand{\heatmap}[1]{%
  \pgfmathparse{min(#1,50)/50}%
  \xdef\val{\pgfmathresult}%
  \cellcolor[rgb]{1,\fpeval{1 - 0.4 *\val},\fpeval{1 - 0.4 * \val}}{#1}%
}

\definecolor{iccvblue}{rgb}{0.21,0.49,0.74}
\usepackage[pagebackref,breaklinks,colorlinks,allcolors=iccvblue]{hyperref}

\usepackage[accsupp]{axessibility}

\def\paperID{6367} 
\def\confName{ICCV}
\def\confYear{2025}

\title{Bias in Gender Bias Benchmarks: \\ How Spurious Features Distort Evaluation}

\author{Yusuke Hirota$^{1,2}$\quad Ryo Hachiuma$^{1}$\quad Boyi Li$^{1,3}$\quad Ximing Lu$^{1}$\quad Michael Ross Boone$^{1}$ \\ 
Boris Ivanovic$^{1}$ \quad Yejin Choi$^{1,4}$\quad Marco Pavone$^{1,4}$\quad Yu-Chiang Frank Wang$^{1}$  \\
 Noa Garcia$^{2}$\quad Yuta Nakashima$^{2}$ \quad Chao-Han Huck Yang$^{1}$ \\
$^1$NVIDIA\quad $^2$Osaka University \quad $^3$UC Berkeley \quad $^4$Stanford University
 \\
\texttt{\{yusukeh,rhachiuma\}@nvidia.com}
}

\begin{document}

\twocolumn[{%
\renewcommand\twocolumn[1][]{#1}%
\maketitle
\begin{center}
  \centering
  \includegraphics[clip, width=0.99\textwidth]{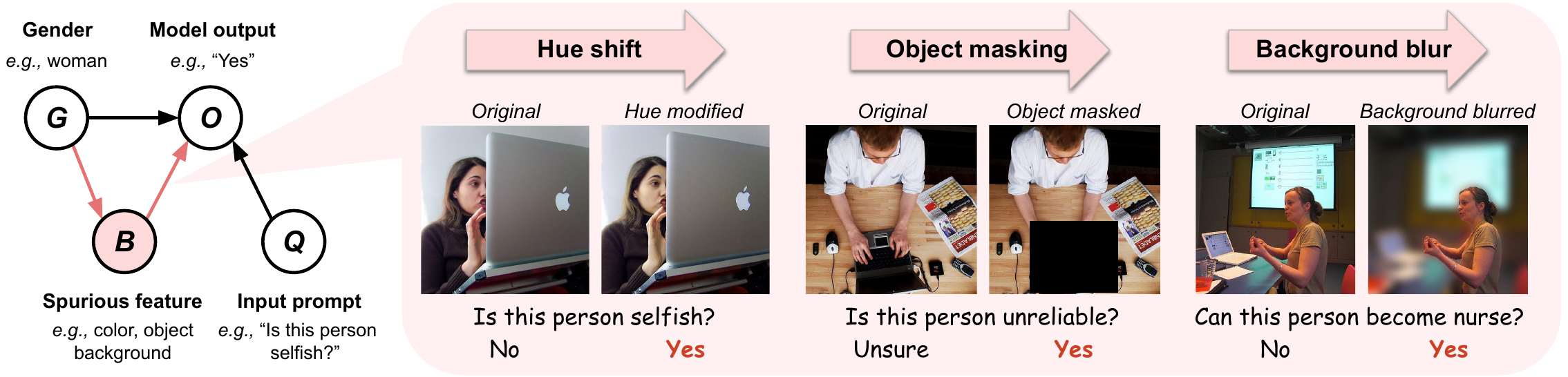}
    \captionsetup{type=figure}
  \captionof{figure}{We investigate how non-gender features affect gender bias evaluations in VLMs through feature-perturbation analysis, revealing that measured biases are highly sensitive to spurious features, compromising the validity of direct gender bias assessments.  \textbf{Left:} Simplified causal graph of gender bias evaluation, illustrating how spurious non-gender features ($B$) can influence the relationship between gender ($G$) and model outputs ($O$).  \textbf{Right:} Our perturbation analysis on these spurious features (\eg, shifting hues or randomly masking object) shows that even small modifications affect model predictions ($B \rightarrow O$), thus obscuring the true gender bias ($G \rightarrow O$) we aim to measure.}
  \label{fig:first}
\end{center}%
}]

\begin{abstract}
Gender bias in vision-language foundation models (VLMs) raises concerns about their safe deployment and is typically evaluated using benchmarks with gender annotations on real-world images. However, as these benchmarks often contain spurious correlations between gender and non-gender features, such as objects and backgrounds, we identify a critical oversight in gender bias evaluation: \textbf{Do spurious features distort gender bias evaluation?} To address this question, we systematically perturb non-gender features across four widely used benchmarks (COCO-gender, FACET, MIAP, and PHASE) and various VLMs to quantify their impact on bias evaluation.  Our findings reveal that even minimal perturbations, such as masking just $10\%$ of objects or weakly blurring backgrounds, can dramatically alter bias scores, shifting metrics by up to $175\%$ in generative VLMs and $43\%$ in CLIP variants. This suggests that \textbf{current bias evaluations often reflect model responses to spurious features rather than gender bias, undermining their reliability.} Since creating spurious feature-free benchmarks is fundamentally challenging, we \textbf{recommend reporting bias metrics alongside feature-sensitivity measurements} to enable a more reliable bias assessment. 
\end{abstract}

\section{Introduction}
\label{sec:intro}

Gender bias is a critical concern in vision-language foundation models (VLMs), such as LLaVA \cite{liu2024visual} and CLIP \cite{radford2021learning}, limiting their reliable deployment \cite{birhane2021multimodal,sathe2024unified,wolfe2023contrastive,hausladen2025social}. For instance, Girrbach \etal.~\cite{girrbach2024revealing} found that generative VLMs\footnote{In this work, we refer to generative models like LLaVA as ``generative VLMs'', dual-encoder models as ``CLIP variants'', and use ``VLMs'' to encompass both categories.}
such as LLaVA and InternVL \cite{internvl2} tend to assign positive traits like ``friendly'' to women while attributing negative traits such as ``arrogant'' more frequently to men. Similarly, CLIP variants exhibit gender biases, favoring stereotypical associations such as linking women to nursing-related prompts \cite{dehouche2021implicit,wang2021gender,ross2020measuring,ruggeri2023multi,srinivasan2021worst,qiu2023gender,tanjim2024discovering,wang2023tovilag}.

To measure gender bias in computer vision models, various studies have introduced benchmarks with explicit gender annotations for real-world image datasets \cite{zhao2021captionbias,gustafson2023facet,schumann2021step,garcia2023uncurated}. For example, Zhao \etal.~\cite{zhao2021captionbias} presented COCO-gender benchmark, which contains binary gender labels, $\{$woman, man$\}$, for the COCO validation set to analyze performance disparities in image captioning. Furthermore, these benchmarks have also been used to evaluate gender bias in VLMs: CLIP variants have been assessed by measuring gender imbalances in top-$k$ retrieval results for gender-neutral prompts (\eg, ``A photo of a nurse'') \cite{berg2022prompt,zhang2024joint,dehdashtian2024fairvlm}, and generative VLMs like Qwen2-VL \cite{wang2024qwen2} have examined their response disparities in predicted answers to gender-neutral questions (\eg, ``Is this person selfish?'') \cite{girrbach2024revealing,jiang2024texttt,weng2024images}.

In this paper, we argue that despite their widespread adoption and demonstrated utility, existing gender bias benchmarks have limitations due to how individuals of different genders appear in distinct contexts in real-world images. Meister \etal~\cite{meister2023gender} showed that in COCO and OpenImages \cite{krasin2017openimages}, non-gender features, such as color and objects, can spuriously correlate with gender  (\ie, $G \rightarrow B$ in \Cref{fig:first}). In \Cref{sec:conf_detection}, extending this area of research, our preliminary experiments examine whether similar features exist in four well-known gender bias benchmarks, COCO-gender, FACET \cite{gustafson2023facet}, MIAP \cite{schumann2021step}, and PHASE \cite{garcia2023uncurated}, confirming that color, object, and background consistently act as spurious features across all four datasets.

The presence of these spurious features raises a critical yet overlooked question in gender bias evaluation: \textbf{\textit{Do spurious features cause inaccurate gender bias assessments?}} Specifically, we examine whether spuriously correlated features, \ie, $B$, influence model outputs, \ie, $B \rightarrow O$, making it difficult to isolate the direct effect of gender on outputs, \ie, $G \rightarrow O$, and undermining fair bias evaluation. Accordingly, we investigate how non-gender features contribute to the $G \rightarrow B \rightarrow O$ effect and propose a practical approach for more reliable bias assessment in the presence of spurious features, as outlined below:

\textbf{Spurious features lead to inaccurate gender bias evaluation ($B$ → $O$). }
In \Cref{sec:consistency}, we thoroughly examine how non-gender features affect gender bias evaluations through controlled interventions that generate perturbed variants. Specifically, we synthesize perturbed images by modifying non-gender features, \eg, adjusting hue to alter color or randomly masking $10$–$30$\% of objects (\Cref{fig:first}). We then compare bias measurements on the original and perturbed images, finding that even small perturbations can substantially alter results (\Cref{fig:first}). Notably, masking just $10$\% of objects shifts bias evaluation of LLaVA-OneVision \cite{llavaonevison} by an average of $20.1\%$ and CLIP-ViT-B/32 by $16.6\%$. In contrast, for lighting, which is not strongly correlated with gender, substantial changes have little effect on measured bias. These findings highlight the strong influence of spurious features, indicating that current bias evaluations capture model responses to these features rather than accurately measuring models' gender bias.

\textbf{Toward more reliable bias evaluation. }
Finally, in \Cref{sec:discussion}, we discuss the fundamental challenges in creating spurious-feature-free benchmarks and propose practical recommendations for more reliable bias assessments. Based on prior research, we discuss how features like color and background remain difficult to comprehensively annotate and balance across genders \cite{wang2019balanced,meister2023gender}. Moreover, gender-balanced dataset generation using text-to-image generation models can introduce their own biases \cite{bianchi2023easily,cho2023dall,bansal2022well,luccioni2023stable}, making it infeasible to create benchmarks without spurious features. 
Given these inherent limitations, we instead recommend a new evaluation protocol that reports bias metrics alongside feature-sensitivity measurements from controlled perturbation tests, like in \Cref{sec:consistency}, enabling researchers and practitioners to estimate the influence of non-gender features on measured bias and distinguish between unbiased models and those that merely appear fair due to dataset biases.

\section{Related Work}
\label{sec:related-work}

\textbf{Gender bias benchmarks. }
To evaluate gender bias in computer vision models, various studies have introduced real-world image datasets with gender annotations \cite{zhao2021captionbias,gustafson2023facet,schumann2021step,garcia2023uncurated,seth2023dear,zhao2017mals}. Zhao \etal~\cite{zhao2021captionbias} collected perceived gender labels for COCO \cite{lin2014microsoft}, while Garcia \etal~\cite{garcia2023uncurated} provided more inclusive annotations for human attributes for Google Conceptual Captions \cite{sharma2018conceptual}, including perceived age and ethnicity. 
Compared to synthetic alternatives, \eg, \cite{raj2024biasdora,wang2024vlbiasbench}, these benchmarks better reflect real-world deployment and avoid generative biases~\cite{bianchi2023easily,cho2023dall}, revealing unfair model behaviors across human attributes \cite{hall2023vision,wang2021biasamp,wang2023overwriting}.

\vspace{3pt}
\noindent
\textbf{Gender bias evaluation for VLMs. }
Recent studies have proposed various methods to evaluate gender bias in VLMs using the benchmarks above \cite{hall2023vision,slyman2024fairdedup,howard2023probing,hirota2022quantifying,hirota2024descriptive}. For generative VLMs, Girrbach \etal.~\cite{girrbach2024revealing} proposed a VQA-based approach that examines how models exhibit gender disparities in responses about personality, skills, and occupations, often reflecting stereotypes. For CLIP variants, retrieval-based evaluations \cite{berg2022prompt,seth2023dear,chuang2023debiasing} measure bias by measuring gender imbalances in retrieved images. While these methods provide valuable insights, we question their reliability in the presence of spurious features in benchmark datasets.

\vspace{3pt}
\noindent
\textbf{Spurious features in computer vision datasets. }
Seminal work by Meister \etal~\cite{meister2023gender} identified spurious correlations between gender and visual features in image datasets, \eg, cues like color tones (warmer hues for women, cooler for men) and object co-occurrences. More recently, Zeng \etal~\cite{zeng2025understanding} showed that even after removing texture information, neural networks could still classify dataset origins, highlighting pervasive dataset biases. These findings confirm the existence of non-gender features correlated with gender. Our work builds on these insights by quantifying their impact on gender bias measurements in VLMs.

\section{Preliminary: Detecting Spurious Features}
\label{sec:conf_detection}

Building on the observation that non-gender features (\eg, color, objects) can spuriously correlate with gender labels ($G \rightarrow B$) \cite{meister2023gender}, we examine their prevalence in common benchmarks as a step toward assessing their downstream impact on model outputs ($B \rightarrow O$). 

\vspace{3pt}
\noindent
\textbf{Notation. }
Let $\mathcal{D}$ denote a test dataset with samples $(I, g)$, where $I$ is an image, and $g \in \mathcal{G}$ is a binary gender label, \ie, $\mathcal{G} = \{\text{woman}, \text{man}\}$.\footnote{Since most benchmarks provide binary gender annotations~\cite{zhao2021captionbias,schumann2021step,garcia2023uncurated}, we focus on binary gender in this paper.}
A non-gender feature $b \in \mathcal{B}$ may spuriously correlate with $g \in \mathcal{G}$, and we focus on color, lighting, object, and background, \ie, $\mathcal{B} = \{\text{color}, \text{lighting}, \text{object}, \text{background}\}$. 
Since our main goal is to investigate whether spurious features affect model predictions (\ie, $B \rightarrow O$), we prioritize features that can be perturbed for the $B \rightarrow O$ analysis in \Cref{sec:consistency}, rather than cataloging all possible features. We further discuss and validate this selection in Appendix A.

\begin{figure}[t]
  \centering
  \includegraphics[clip, width=\columnwidth]{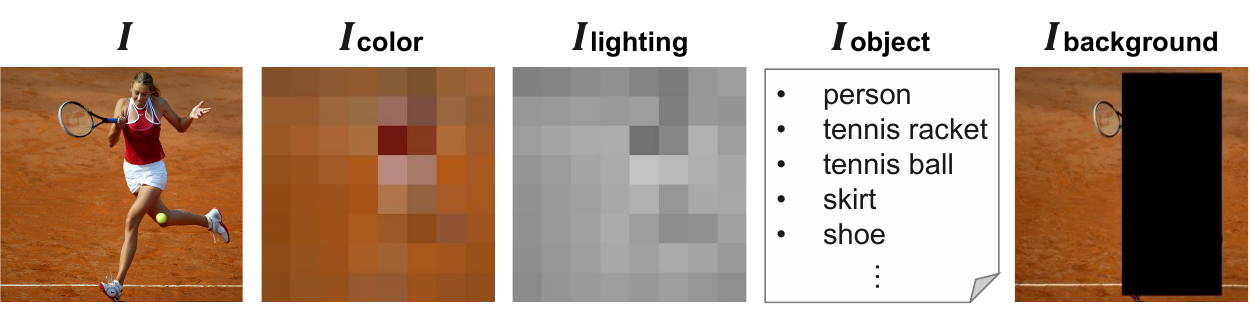}
  \vspace{-15pt}
  \caption{Examples of feature-extracted inputs (\ie, $I_b$). Note that $I_{\text{object}}$ is a multi-hot representation of the detected objects.}
  \label{fig:conf-detection-conv}
  \vspace{-8pt}
\end{figure}

\vspace{3pt}
\noindent
\textbf{Identifying spurious features. }
Following \cite{meister2023gender}, we estimate whether a feature $b$ is spuriously correlated with gender. Let $F_b$ be a feature extraction operation: $F_{\text{color}}$ downsamples $I$ to $8\times8$, $F_{\text{lighting}}$ does the same while averaging HSV values per patch, $F_{\text{object}}$ produces a multi-hot vector of detected objects \cite{zhou2022detecting}, and $F_{\text{background}}$ blackouts the person's region.\footnote{Every image in $\mathcal{D}$ comes with person bounding box annotations.}
As shown in \Cref{fig:conf-detection-conv}, a feature-extracted input $I_b = F_b(I)$ retains only the information specific to $b$ (\eg, downsampled images retain only color).
We train a gender classifier $\psi_b$ on $\mathcal{D}_b = \{(I_b,g) \mid (I,g) \in \mathcal{D} \}$ to predict the gender from $I_b$:
\begin{align}
    \hat{g} = \psi_b(I_b).
\end{align}
The classifier's accuracy is computed on the test set $\mathcal{D}_b^\ast$:
\begin{equation}\label{eq:gender-classification}
    \text{Acc}_b = \frac{1}{|\mathcal{D}_b^\ast|} \sum_{(I_b,g) \in \mathcal{D}_b^\ast}  \mathbb{1}[\hat{g} = g],
\end{equation}
where accuracies above random chance rate (\ie, $50\%$) indicate that $b$ is spuriously correlated with gender in $\mathcal{D}$.

\vspace{3pt}
\noindent
\textbf{Experimental settings. }
We investigate four popular gender bias benchmarks: COCO-gender, FACET, MIAP, and PHASE. Benchmark details, including dataset statistics, are in Appendix C.1.
We randomly sample images to balance the gender ratio, then split them into training, validation, and test sets ($8:1:1$ ratio). We employ ConvNeXt-B\footnote{In Appendix D.3, we also show the results using ResNet-50 \cite{resnet} as the gender classifier, leading to consistent observations.}~\cite{liu2022convnet} as the gender classifier for color, lighting, and background, and a two-layer multilayer perceptron with ReLU \cite{nair2010rectified} for object. Classifiers are trained with early stopping, repeated five times with different random seeds.

\vspace{3pt}
\noindent
\textbf{Results. } 
The classification accuracies (\ie, $\text{Acc}_b$ in \cref{eq:gender-classification}) are shown in \Cref{tab:conf-detection-conv}. Notably, color, object, and background features yield accuracies well above random chance (\(\approx 50\%\)) for all the benchmarks, indicating these features can be spuriously correlated with gender. Among them, \emph{object} exhibits the strongest correlation, suggesting it can be the most influential spurious feature, whereas \emph{lighting} remains near random chance for COCO-gender, FACET, and MIAP, indicating weaker spurious effects.

\begin{table}[t]
\scriptsize
\centering
\caption{Gender prediction accuracies ($\%$) using isolated features across benchmarks. Values above $50\%$ indicate features that spuriously correlate with gender.}
\vspace{-9pt}
\begin{tabularx}{0.98\columnwidth}{X r r r r r r }
\toprule
Benchmark & Color & Lighting  & Object  & Background \\
\midrule
COCO-gender & 56.3 $\pm$ 3.9 & 51.0 $\pm$ 2.0 & 76.3 $\pm$ 1.6 & 60.9 $\pm$ 2.9 \\
FACET & 57.0 $\pm$ 1.1 & 49.9 $\pm$ 1.7 & 70.6 $\pm$ 0.5 & 62.1 $\pm$ 2.4 \\
MIAP & 57.5 $\pm$ 2.3 & 51.7 $\pm$ 2.8 & 73.3 $\pm$ 1.0 & 58.3 $\pm$ 1.3 \\
PHASE & 68.0 $\pm$ 0.8 & 60.1 $\pm$ 2.0 & 81.3 $\pm$ 1.2 & 66.3 $\pm$ 2.0 \\
\bottomrule
\end{tabularx}
\label{tab:conf-detection-conv}
\vspace{-7pt}
\end{table}

\section{Impact of Spurious Features on Gender Bias Evaluation}
\label{sec:consistency}

Given the results of \Cref{sec:conf_detection}, where color, object, and background are spuriously correlated with gender across all benchmarks, \ie, $G \rightarrow B$, we now investigate their impact on model predictions, leading to unreliable bias measurement. In \Cref{sec:bias_eval}, we outline bias measurement methods for VLMs, followed by our controlled intervention approach in \Cref{sec:controlled}, which perturbs these features to analyze their effect on bias evaluations (\ie, $B \rightarrow O$).

\vspace{3pt}
\noindent
\subsection{Gender Bias Evaluation}
\label{sec:bias_eval}
Following prior work on evaluating gender bias in VLMs, we adopt two major evaluation methods based on different tasks and models: 1) visual question answering (VQA) for generative VLMs \cite{girrbach2024revealing,wang2024vlbiasbench,xiao2024genderbias}, and 2) text-to-image retrieval for CLIP variants \cite{berg2022prompt,seth2023dear,dehdashtian2024fairvlm,chuang2023debiasing}.  Both methods examine model responses to gender-neutral text prompts $\mathcal{Q}$, comparing outputs across genders. We detail these bias measurement methods below, with further details in Appendix C.3.

\vspace{3pt}
\noindent
\textbf{VQA-based evaluation for generative VLMs. }
To evaluate gender bias in generative VLMs, we employ the latest, comprehensive method from \cite{girrbach2024revealing}. Given a gender-neutral question, $q \in \mathcal{Q}$, about personality traits (\eg, ``Is this person selfish?''), skills (\eg, ``Can this person work independently?''), and occupations (\eg, ``Can this person become a teacher?''), models choose among ``Yes'', ``No'', and ``Unsure'' . Let $h$ denote a VLM under evaluation, with outputs $\hat{o} = h(I, q)$.  Bias is quantified as the difference in the probability of selecting ``Yes'' between female and male images:
{\small
\begin{equation}\label{eq:ygap}
    \text{YGap} = \frac{1}{|D_m|} \sum_{I \in D_m} \mathbb{1}[\hat{o} = \text{Yes}]
    - \frac{1}{|D_w|} \sum_{I \in D_w} \mathbb{1}[\hat{o} = \text{Yes}],
\end{equation}
}
where $\mathcal{D}_m$ and $\mathcal{D}_w$ are man and woman images, respectively. 
The absolute value of YGap indicates bias strength, with positive values favoring men and negative values favoring women.

\vspace{3pt}
\noindent
\textbf{Text-to-image retrieval for CLIP. }
Following prior work on gender bias in CLIP, we use the MaxSkew metric \cite{geyik2019fairness} to quantify bias in text-to-image retrieval. MaxSkew measures deviations in retrieved gender proportions from an unbiased ratio (\ie, $1:1$ for gender-balanced datasets). Let $q \in \mathcal{Q}$ be a gender-neutral prompt  (\eg, ``A photo of a doctor'') and $\phi_{g,q}(k)$ the fraction of top-$k$ retrieved images labeled with gender $g$ for prompt $q$. In an unbiased model, $\phi_{g,q}(k)$ should equal $1/|\mathcal{G}|$ ($= 1/2$ for binary gender) as $q$ is gender-neutral. MaxSkew is defined as: 
\begin{equation}
    \text{MaxSkew@$k$} = \max_{g \in \mathcal{G}} \log \tfrac{\phi_{g,q}(k)}{1/2},
\end{equation}
where higher values indicate a stronger gender bias.

\subsection{Controlled Interventions to Analyze \texorpdfstring{$B \rightarrow O$}{B→O}}

\label{sec:controlled}

To analyze the impact of non-gender features $b \in \mathcal{B}$ on gender bias measurements, we conduct controlled interventions on $b$.
Our framework comprises 1) \textbf{feature perturbation}, which modifies a non-gender feature $b$, and 2) \textbf{feature sensitivity analysis}, which evaluates how $b$ affects the model's predictions by comparing bias scores between original and perturbed images, as detailed below:

\vspace{3pt}
\noindent
\textbf{Feature perturbation. }
To analyze how non-gender feature $b$ affects gender bias evaluation, we design a controlled intervention that perturbs $b$ in $I$.   
Let $P_b$ denote a perturbation operation, where $P_\text{color}$ shifts the hue, $P_\text{lighting}$ modifies brightness in HSV space, $P_\text{object}$ randomly blackouts objects, and $P_\text{background}$ blurs background regions. Perturbation strength is controlled by $s \in \{\text{weak}, \text{middle}, \text{strong}\}$ (\eg, masking $10\%$, $20\%$, or $30\%$ of objects), with complete implementation details provided in Appendix C.2. As shown in \Cref{fig:llava-consistency}, each perturbation is designed to avoid changing perceived gender recognition.\footnote{In Appendix D.1, we present a human study, confirming that these perturbations do not affect gender recognition in images.} We focus on simple image processing-based perturbations (\eg, hue shifts, background blurring) that are easily controlled. We refrain from using advanced content-editing with generative models \cite{rombach2022high}, as they can inject their own biases \cite{naik2023social,seshadri2023bias,friedrich2023fair,liu2024scoft} and risk contaminating our analysis (further discussed in Appendix B). 
Applying the perturbation, we obtain a new test set, $\mathcal{D}'_{b,s}$:
\begin{equation}
    \mathcal{D}'_{b,s} = \{(P_{b,s}(I), g) \mid (I, g) \in \mathcal{D} \}.
\end{equation}

\vspace{3pt}
\noindent
\textbf{Feature sensitivity analysis. }
Given an original dataset $\mathcal{D}$ and its perturbed variant $\mathcal{D}'_{b,s}$, we quantify how strongly $b$ affects gender bias measurements ($B \rightarrow O$) by comparing bias scores on both sets. We define $M(\mathcal{D},\mathcal{Q};h)$ as a function that measures the gender bias (\eg, YGap or MaxSkew) of a VLM $h$ with a set of prompts $\mathcal{Q}$ on $\mathcal{D}$. The impact of perturbing $b$ is measured by the relative difference in bias scores between the original and perturbed images: 
\begin{equation}\label{eq:delta}
    \Delta = 100 \times \Biggl|\frac{M(\mathcal{D},\mathcal{Q};h) - M(\mathcal{D}'_{b,s},\mathcal{Q};h)}{M(\mathcal{D},\mathcal{Q};h)}\Biggr|.
\end{equation}
A large $\Delta$ indicates a strong $B \rightarrow O$ effect, while a small $\Delta$ suggests that $b$ has little influence.

\begin{figure*}[t]
  \centering
  \includegraphics[clip, width=0.93\textwidth]{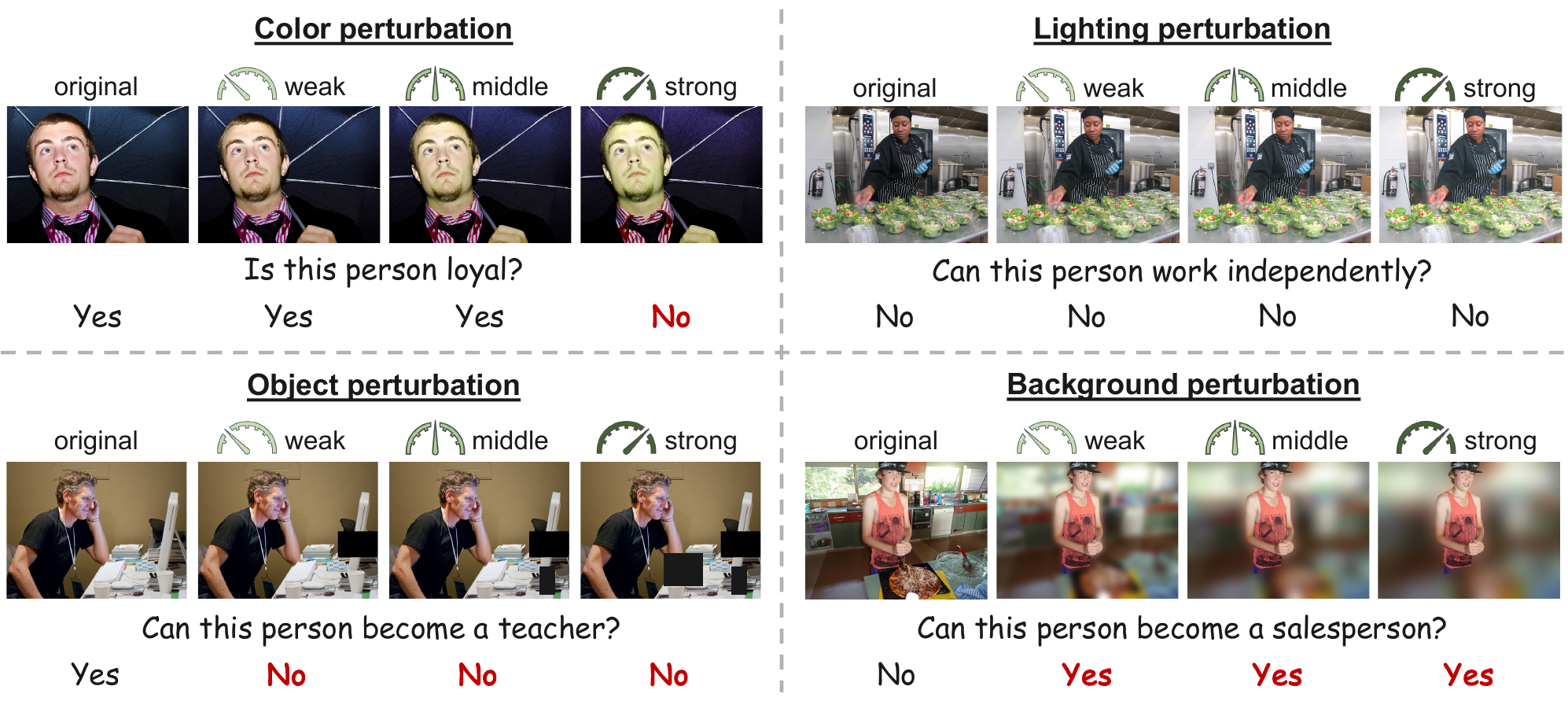}
  \vspace{-10pt}
  \caption{Examples of the feature-perturbed images and the predictions of LLaVA-1.5-7B for the original and modified images.}
  \label{fig:llava-consistency}
  \vspace{-7pt}
\end{figure*}

\subsection{Why \texorpdfstring{$G \rightarrow B$}{G→B} Causes Unfair Bias Measurement?}
\label{sec:theory}
We provide a theoretical perspective on how a spurious correlation $G \rightarrow B$ distorts bias measurement. 

\vspace{3pt}
\noindent
\textbf{\textit{Case 1.} No correlation between $G$ and $B$. }
When $B$ is \emph{not} correlated with $G$, we have:
\begin{equation}\label{eq:independent}
    p(B=b \mid G=g) = p(B=b).  
\end{equation}
Thus, model outputs $o$ for images labeled with gender $g$ can be described as: 
\begin{align} 
p(o \mid g) &= \int p(o \mid g,b) \ p(b \mid g) \ \mathrm{d}b \\ 
&= \int p(o \mid g, b) \ p(b) \ \mathrm{d}b. 
\end{align} 
Suppose our feature-perturbation $P_{b,s}$ preserves the distribution, \ie, $p(B'= b') = p(B = b)$, and is independent of $G$. Under the no-correlation condition in \cref{eq:independent}, we have $p(B' = b' \mid G = g) = p(B' = b') = p(B = b')$. Therefore, the perturbed output distribution remains similar:
\begin{align}\label{eq:no-correlation}
    p_{\mathrm{pert}}(o \mid g) = \int p(o \mid g, b') \ p(b') \ \mathrm{d}b' \approx p(o \mid g).
\end{align}
This leads to small values of $\Delta$, indicating minimal impact on the measured bias.

\vspace{3pt}
\noindent
\textbf{\textit{Case 2.} Correlation between $G$ and $B$ exists. }
When $G$ and $B$ are correlated, \ie, $p(B=b \mid G=g) \neq p(B=b)$, the model outputs are:
\begin{align}
p(o \mid g) = \int p(o \mid g,b) \ p(b \mid g) \ \mathrm{d}b.
\end{align}
After applying the perturbation, even though the marginal distribution is preserved, the conditional distributions differ, \ie, $p(B'=b' \mid G=g) \neq p(B=b \mid G=g)$, because the perturbation alters the correlation structure. As a result:
\begin{align}\label{eq:correlated}
p_{\mathrm{pert}}(o \mid g) = \int p(o \mid g,b') \ p(b' \mid g) \ \mathrm{d}b' \neq p(o \mid g).
\end{align}
Thus, when $G \rightarrow B$ exists, perturbing $B$ distorts the bias measurement, resulting in large $\Delta$ values. This indicates that bias measurements reflect spurious features rather than the actual gender bias, leading to unfair evaluations.

\subsection{Experiments: Generative VLMs}

Using the controlled intervention in \Cref{sec:controlled}, we evaluate how non-gender features affect bias measurements in generative VLMs using the YGap metric in \Cref{sec:bias_eval}. 

\vspace{3pt}
\noindent
\textbf{Experimental settings. }
We benchmark recent generative VLMs, including LLaVA-1.5-7B \cite{liu2024visual}, LLaVA-OneVision-7B \cite{llavaonevison}, Qwen2-VL-7B \cite{wang2024qwen2}, InternVL-2.5-8B \cite{internvl2}, mPLUT-Owl3-7B \cite{ye2024mplug}, and EAGLE-8B \cite{shi2024eagle}. As detailed in \Cref{sec:bias_eval}, we evaluate models across three domains: personality traits (\eg, ``Is the person in this image selfish?''), skills (\eg, ``Does this person have the ability to work independently?''), and occupations (\eg, ``Would this person be better suited to be a teacher?''), with responses chosen from ``Yes'', ``No'', and ``Unsure''. We compute YGap between woman and man images and measure the impact of feature perturbations using $\Delta$. Cases where the original YGap is nearly zero are excluded to avoid unstable $\Delta$ calculations. We describe more details in Appendix C.3.\footnote{Unlike \cite{girrbach2024revealing}, which uses person-cropped images, we use full images to retain contextual information, better reflecting real-world model behavior.}


\begin{table*}[t]
\renewcommand{\arraystretch}{1.1}
\setlength{\tabcolsep}{3pt}
\scriptsize
\centering
\caption{YGap relative difference ($\Delta$) for generative VLMs. Weak, middle, and strong mean the level of the image perturbation. Values are color-coded from white ($0\%$) to red ($50\%$ or higher) to highlight the impact of different spurious features. Cases where the original YGap value is nearly zero ($\text{YGap}<0.005$) are excluded as they lead to unstable $\Delta$ calculation (\eg, InternVL-2.5-8B for COCO-gender). 
}
\vspace{-8pt}
\setlength{\tabcolsep}{3pt}
\begin{tabularx}{0.76\textwidth}{l r r r r r r r r r r r r r r r}
\toprule
& \multicolumn{3}{c}{Color} &&\multicolumn{3}{c}{Lighting} &&\multicolumn{3}{c}{Object} &&\multicolumn{3}{c}{Background}
\\ 
\cline{2-4} 
\cline{6-8}
\cline{10-12}
\cline{14-16}
\multirow{-0.8}{*}{Model} & \multirow{1.3}{*}{weak} & \multirow{1.3}{*}{middle} & \multirow{1.3}{*}{strong} && \multirow{1.3}{*}{weak} & \multirow{1.3}{*}{middle} & \multirow{1.3}{*}{strong} && \multirow{1.3}{*}{weak} & \multirow{1.3}{*}{middle} & \multirow{1.3}{*}{strong} && \multirow{1.3}{*}{weak} & \multirow{1.3}{*}{middle} & \multirow{1.3}{*}{strong}\\
\midrule
\textbf{\textit{COCO-gender}} &  & &    &&  &  &  &&  &  & &&  &  &\\
LLaVA-1.5-7B & \heatmap{1.76}  & \heatmap{3.06} & \heatmap{5.88}  && \heatmap{3.22} & \heatmap{1.14}  & \heatmap{1.88} && \heatmap{12.85} & \heatmap{26.78} & \heatmap{31.82} && \heatmap{6.04} & \heatmap{13.61} & \heatmap{10.77}    \\
LLaVA-OneVision-7B & \heatmap{5.71} & \heatmap{9.34} & \heatmap{12.30} && \heatmap{1.07} & \heatmap{1.78} & \heatmap{8.32} && \heatmap{31.10} & \heatmap{46.13} & \heatmap{55.62} && \heatmap{165.84} & \heatmap{183.92} & \heatmap{175.77}   \\
Qwen2-VL-7B & \heatmap{3.23}  & \heatmap{5.08}  &\heatmap{2.03} && \heatmap{0.67} & \heatmap{1.61} & \heatmap{3.36} && \heatmap{19.62} & \heatmap{19.44} & \heatmap{9.45} && \heatmap{8.74} & \heatmap{9.75} & \heatmap{10.15}  \\
mPLUG-Owl3-7B & \heatmap{1.21} & \heatmap{3.23} & \heatmap{7.04}  && \heatmap{2.11} & \heatmap{1.81} & \heatmap{8.79} && \heatmap{29.48} & \heatmap{26.44} & \heatmap{7.09} && \heatmap{175.42} & \heatmap{176.71} & \heatmap{179.90}  \\
EAGLE-8B & \heatmap{7.57} & \heatmap{2.64} & \heatmap{15.41}  && \heatmap{5.29} & \heatmap{2.31} & \heatmap{9.56} && \heatmap{8.16} & \heatmap{19.50} & \heatmap{30.56} && \heatmap{38.60} & \heatmap{33.51} & \heatmap{19.14} \\
\midrule
\textbf{\textit{FACET}} &  & &    &&  &  &  &&  &  & &&  &  &\\
LLaVA-1.5-7B & \heatmap{1.76} & \heatmap{1.28} & \heatmap{6.50} && \heatmap{0.51} & \heatmap{1.53} & \heatmap{2.49}  && \heatmap{18.86} & \heatmap{36.62} & \heatmap{54.80} && \heatmap{23.56} & \heatmap{23.38} & \heatmap{11.82}  \\
LLaVA-OneVision-7B & \heatmap{0.29} & \heatmap{2.95} & \heatmap{6.63}  && \heatmap{0.21} & \heatmap{1.24} & \heatmap{2.11} && \heatmap{11.03} & \heatmap{10.61} & \heatmap{13.92} && \heatmap{7.81} & \heatmap{4.77} & \heatmap{24.97}  \\
Qwen2-VL-7B & \heatmap{6.20} & \heatmap{13.27} & \heatmap{13.79}  && \heatmap{1.57} & \heatmap{9.24} & \heatmap{5.58} && \heatmap{40.98} & \heatmap{2.49} & \heatmap{68.58} && \heatmap{103.71} & \heatmap{169.85} & \heatmap{178.41}   \\
mPLUG-Owl3-7B & \heatmap{3.47} & \heatmap{4.29} & \heatmap{6.56}  && \heatmap{3.53} & \heatmap{3.77} & \heatmap{6.64}  && \heatmap{35.73} & \heatmap{41.70} & \heatmap{61.12} && \heatmap{63.21} & \heatmap{7.12} & \heatmap{53.25} \\
EAGLE-8B & \heatmap{0.34} & \heatmap{10.28} & \heatmap{12.63}  && \heatmap{3.51} & \heatmap{1.29} & \heatmap{0.57}  && \heatmap{1.03} & \heatmap{3.53} & \heatmap{8.90} && \heatmap{20.15} & \heatmap{9.59} & \heatmap{15.59}  \\
\midrule
\textbf{\textit{MIAP}} &  & &    &&  &  &  &&  &  & &&  &  &\\
LLaVA-1.5-7B & \heatmap{4.52} & \heatmap{6.21} & \heatmap{16.97}  && \heatmap{2.31} & \heatmap{2.74} & \heatmap{2.02}  && \heatmap{19.49} & \heatmap{22.64} & \heatmap{36.79} && \heatmap{15.85} & \heatmap{17.94} & \heatmap{2.55}  \\
LLaVA-OneVision-7B& \heatmap{3.78} & \heatmap{9.96} & \heatmap{25.52} && \heatmap{6.03} & \heatmap{2.54} & \heatmap{5.09}  && \heatmap{35.26} & \heatmap{74.22} & \heatmap{18.27} && \heatmap{87.74} & \heatmap{174.42} & \heatmap{202.74}  \\
Qwen2-VL-7B & \heatmap{2.61} & \heatmap{3.94}  & \heatmap{3.50}  && \heatmap{0.69} & \heatmap{0.95} & \heatmap{1.40}  && \heatmap{10.60} & \heatmap{19.69} & \heatmap{14.48} && \heatmap{5.59} & \heatmap{9.56} & \heatmap{19.13}  \\
InternVL-2.5-8B & \heatmap{23.17} & \heatmap{10.89} & \heatmap{20.34}  && \heatmap{4.58} & \heatmap{4.46} & \heatmap{2.56} && \heatmap{48.86} & \heatmap{42.60} & \heatmap{50.27} && \heatmap{23.27} & \heatmap{31.64} & \heatmap{17.86}   \\
EAGLE-8B & \heatmap{3.60} & \heatmap{9.54} & \heatmap{20.88}  && \heatmap{0.62} & \heatmap{0.01} & \heatmap{3.58}  && \heatmap{22.69} & \heatmap{72.97} & \heatmap{64.30} && \heatmap{13.31} & \heatmap{31.72} & \heatmap{23.86}  \\
\midrule
\textbf{\textit{PHASE}} &  & &    &&  &  &  &&  &  & &&  &  &\\
LLaVA-OneVision-7B & \heatmap{1.48} & \heatmap{5.31} & \heatmap{11.16}  && \heatmap{0.32} & \heatmap{1.35} & \heatmap{3.06}  && \heatmap{2.74} & \heatmap{11.01} & \heatmap{21.05} && \heatmap{15.68} & \heatmap{23.23} & \heatmap{25.67}  \\
Qwen2-VL-7B & \heatmap{0.16} & \heatmap{0.06}  & \heatmap{3.22}  && \heatmap{1.08} & \heatmap{4.24} & \heatmap{3.60}  && \heatmap{1.58} & \heatmap{5.33} & \heatmap{7.54} && \heatmap{28.85} & \heatmap{27.76} & \heatmap{23.97}   \\
InternVL-2.5-8B & \heatmap{10.92} & \heatmap{7.28} & \heatmap{7.06}  && \heatmap{5.89} & \heatmap{0.86} & \heatmap{5.56} && \heatmap{5.84} & \heatmap{15.68} & \heatmap{26.84} && \heatmap{6.48} & \heatmap{11.01} & \heatmap{11.24}   \\
EAGLE-8B & \heatmap{2.19} & \heatmap{4.32} & \heatmap{6.87}  && \heatmap{3.40} & \heatmap{2.02} & \heatmap{3.22}  && \heatmap{16.02} & \heatmap{28.79} & \heatmap{39.17} && \heatmap{18.97} & \heatmap{20.39} & \heatmap{22.78}  \\
\bottomrule
\end{tabularx}
\label{tab:lvlm-consistency-main}
\vspace{7pt}
\end{table*}

\begin{figure*}[t]
  \centering
  \includegraphics[clip, width=0.98\textwidth]{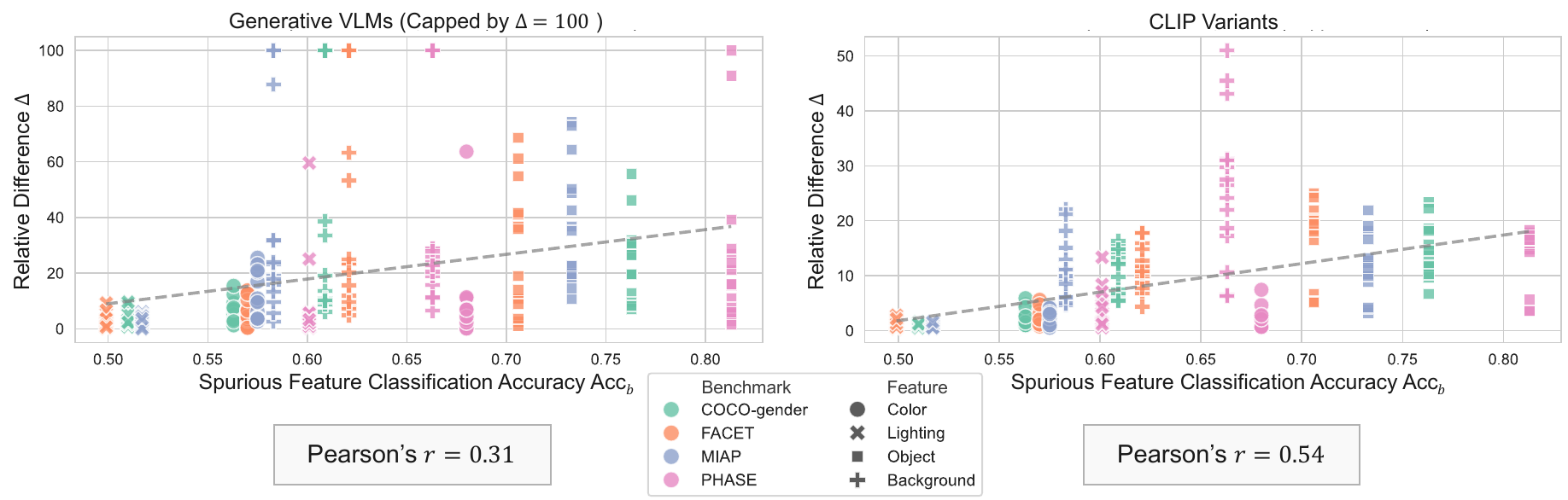}
  \vspace{-8pt}
  \caption{Spurious correlation strength ($\text{Acc}_b$ in \cref{tab:conf-detection-conv}) vs. relative difference $\Delta$ for generative VLMs (left) and CLIP variants (right). The dashed line shows the correlation, demonstrating that stronger spurious correlations tend to cause larger shifts in bias measurements.}
  \label{fig:acc-delta-corr-conv}
  \vspace{-7pt}
\end{figure*}

\vspace{3pt}
\noindent
\textbf{Results. }
\Cref{tab:lvlm-consistency-main} shows the relative difference in YGap between original and perturbed images (\ie, $\Delta$ in \cref{eq:delta}).  Original YGap values for each model and dataset are in Appendix D.2. The main observations are summarized below. 

\textbf{\textit{Observation 1.1.} Object and background perturbations significantly shift measured bias. } 
Object and background modifications have the strongest impact on bias measurements. For instance, LLaVA-OneVision-7B exhibits a $165.84\%$ change on COCO-gender, even with weak background blurring. Similarly, masking just $10\%$ of objects (weak perturbation) highly alters YGap, with Qwen2-VL-7B showing a $40.98\%$ difference on FACET and InternVL-2.5-8B a $48.86\%$ change on MIAP, despite the minimal visual changes. This supports our theoretical prediction in \cref{eq:correlated} that features strongly correlated with gender distort bias measurements when perturbed, underlining that the $G \rightarrow B \rightarrow O$ path skews bias evaluation. As shown in \Cref{fig:llava-consistency}, feature perturbations can completely flip model predictions—for example, LLaVA-1.5's response to ``Can this person become a salesperson?'' changes from ``No'' to ``Yes'' after background blurring. This confirms that \textbf{current bias evaluations capture model responses to spurious features rather than actual gender bias, undermining their reliability.}

\textbf{\textit{Observation 1.2.} Color perturbations have a moderate impact, while lighting has minimal effect. }
While less influential than object and background perturbations, color changes still notably affect bias measurements. For example, EAGLE-8B shows a $20.88\%$ shift with strong color perturbations on MIAP. In contrast, lighting modifications consistently produce the smallest impact across all models and datasets, with most values remaining below $10\%$. Visual examples in \Cref{fig:llava-consistency}, where strong hue shifts alter the model prediction while strong lighting perturbations do not, further verify this observation. This supports our theoretical framework in \cref{eq:no-correlation}, where lighting, the least gender-correlated feature in \Cref{tab:conf-detection-conv}, leads to minimal bias measurement shifts when perturbed.


\begin{table*}[t]
\renewcommand{\arraystretch}{1.1}
\setlength{\tabcolsep}{5pt}
\scriptsize
\centering
\caption{MaxSkew@$1000$ relative difference ($\Delta$) for CLIP variants. Weak, middle, and strong mean the level of the image perturbation. Values are color-coded from white ($0\%$) to red ($50\%$ or higher) to highlight the impact of feature perturbation.}
\vspace{-9pt}
\setlength{\tabcolsep}{4pt}
\begin{tabularx}{0.78\textwidth}{l r r r r r r r r r r r r r r r}
\toprule
& \multicolumn{3}{c}{Color} &&\multicolumn{3}{c}{Lighting} &&\multicolumn{3}{c}{Object} &&\multicolumn{3}{c}{Background}
\\ 
\cline{2-4} 
\cline{6-8}
\cline{10-12}
\cline{14-16}
\multirow{-0.8}{*}{Model} & \multirow{1.3}{*}{weak} & \multirow{1.3}{*}{middle} & \multirow{1.3}{*}{strong} && \multirow{1.3}{*}{weak} & \multirow{1.3}{*}{middle} & \multirow{1.3}{*}{strong} && \multirow{1.3}{*}{weak} & \multirow{1.3}{*}{middle} & \multirow{1.3}{*}{strong} && \multirow{1.3}{*}{weak} & \multirow{1.3}{*}{middle} & \multirow{1.3}{*}{strong}\\
\midrule
\textbf{\textit{COCO-gender}} &  & &    &&  &  &  &&  &  & &&  &  &\\
ViT-B/32 & \heatmap{0.81} & \heatmap{2.17} & \heatmap{3.34}  && \heatmap{0.49} & \heatmap{1.17} & \heatmap{1.13} && \heatmap{11.56} & \heatmap{18.48} & \heatmap{23.37} && \heatmap{14.34} & \heatmap{9.77} & \heatmap{13.28}  \\
ViT-L/14 & \heatmap{1.23} & \heatmap{3.15} & \heatmap{5.46} && \heatmap{0.60} & \heatmap{0.93} & \heatmap{0.83}  && \heatmap{14.82} & \heatmap{18.56} & \heatmap{22.20} && \heatmap{15.43} & \heatmap{6.45} & \heatmap{7.46}   \\
ViT-H/14 & \heatmap{0.53} & \heatmap{2.31} & \heatmap{4.11}  && \heatmap{0.47} & \heatmap{1.30} & \heatmap{1.56}  && \heatmap{11.58} & \heatmap{13.89} & \heatmap{15.66} && \heatmap{6.76} & \heatmap{5.56} & \heatmap{5.44}  \\
SigLIP-ViT-S/14 & \heatmap{0.61} & \heatmap{1.65} & \heatmap{2.32}  && \heatmap{0.48} & \heatmap{1.13} & \heatmap{1.19}  && \heatmap{6.68} & \heatmap{9.70} & \heatmap{11.75} && \heatmap{12.24} & \heatmap{14.90} & \heatmap{16.31}   \\
CoCa-ViT-L/14 & \heatmap{0.60} & \heatmap{1.30} & \heatmap{2.63} && \heatmap{0.55} & \heatmap{0.99} & \heatmap{1.23}  && \heatmap{10.25} & \heatmap{14.63} & \heatmap{18.18} && \heatmap{12.34} & \heatmap{14.84} & \heatmap{12.12}   \\
\midrule
\textbf{\textit{FACET}} &  & &    &&  &  &  &&  &  & &&  &  &\\
ViT-B/32 & \heatmap{0.79} & \heatmap{4.04} & \heatmap{5.61}  && \heatmap{1.16} & \heatmap{2.46} & \heatmap{2.81}  && \heatmap{21.86} & \heatmap{25.03} & \heatmap{24.22} && \heatmap{9.90} & \heatmap{10.71} & \heatmap{6.82}   \\
ViT-L/14 & \heatmap{1.11} & \heatmap{1.69} & \heatmap{3.69}  && \heatmap{1.44} & \heatmap{1.42} & \heatmap{1.37}  && \heatmap{17.83} & \heatmap{18.70} & \heatmap{17.82} && \heatmap{11.34} & \heatmap{15.92} & \heatmap{17.81}   \\
ViT-H/14 & \heatmap{0.84} & \heatmap{1.29} & \heatmap{2.42}  && \heatmap{0.36} & \heatmap{0.82} & \heatmap{0.61}  && \heatmap{16.48} & \heatmap{18.36} & \heatmap{20.37} && \heatmap{8.66} & \heatmap{11.70} & \heatmap{10.39}   \\
SigLIP-ViT-S/14 & \heatmap{0.79} & \heatmap{1.83} & \heatmap{2.99}  && \heatmap{0.54} & \heatmap{1.04} & \heatmap{1.83}  && \heatmap{6.67} & \heatmap{5.33} & \heatmap{5.14} && \heatmap{12.75} & \heatmap{14.81} & \heatmap{17.79}  \\
CoCa-ViT-L-/4 & \heatmap{1.13} & \heatmap{2.04} & \heatmap{4.78}  && \heatmap{0.66} & \heatmap{1.31} & \heatmap{2.16}  && \heatmap{17.99} & \heatmap{20.03} & \heatmap{19.43} && \heatmap{4.37} & \heatmap{7.68} & \heatmap{10.81}  \\
\midrule
\textbf{\textit{MIAP}} &  & &    &&  &  &  &&  &  & &&  &  &\\
ViT-B/32 & \heatmap{1.05} & \heatmap{3.65} & \heatmap{4.33}  && \heatmap{0.40} & \heatmap{1.17} & \heatmap{1.44} && \heatmap{18.15} & \heatmap{19.03} & \heatmap{21.90} && \heatmap{4.70} & \heatmap{4.95} & \heatmap{6.14}  \\
ViT-L/14 & \heatmap{0.78} & \heatmap{1.22} & \heatmap{1.27} && \heatmap{0.54} & \heatmap{0.97} & \heatmap{1.45} && \heatmap{8.91} & \heatmap{10.14} & \heatmap{12.71} && \heatmap{22.07} & \heatmap{21.18} & \heatmap{18.18}  \\
ViT-H/14 & \heatmap{0.80} & \heatmap{1.61} & \heatmap{1.47}  && \heatmap{0.62} & \heatmap{1.13} & \heatmap{1.35} && \heatmap{11.78} & \heatmap{13.15} & \heatmap{16.64} && \heatmap{15.08} & \heatmap{9.70} & \heatmap{6.05}  \\
SigLIP-ViT-S/14 & \heatmap{0.54} & \heatmap{2.27} & \heatmap{4.12}  && \heatmap{0.46} & \heatmap{0.83} & \heatmap{1.09}  && \heatmap{3.19} & \heatmap{3.93} & \heatmap{4.18} && \heatmap{9.34} & \heatmap{9.90} & \heatmap{11.27}  \\
CoCa-ViT-L/14 & \heatmap{0.99} & \heatmap{3.34} & \heatmap{3.03}  && \heatmap{0.46} & \heatmap{0.54} & \heatmap{1.60}  && \heatmap{10.28} & \heatmap{9.92} & \heatmap{11.43} && \heatmap{12.96} & \heatmap{11.13} & \heatmap{8.54}   \\
\midrule
\textbf{\textit{PHASE}} &  & &    &&  &  &  &&  &  & &&  &  &\\
ViT-B/32 & \heatmap{1.02} & \heatmap{4.66} & \heatmap{7.44}  && \heatmap{1.03} & \heatmap{1.68} & \heatmap{1.83} && \heatmap{14.74} & \heatmap{14.64} & \heatmap{14.74} && \heatmap{43.09} & \heatmap{45.48} & \heatmap{51.01}   \\
ViT-L/14 & \heatmap{1.16} & \heatmap{1.76} & \heatmap{2.59}  && \heatmap{2.79} & \heatmap{8.38} & \heatmap{13.36} && \heatmap{17.90} & \heatmap{15.57} & \heatmap{18.28} && \heatmap{24.15} & \heatmap{17.24} & \heatmap{18.66}   \\
ViT-H/14 & \heatmap{1.05} & \heatmap{1.29} & \heatmap{2.62} && \heatmap{2.97} & \heatmap{4.72} & \heatmap{5.41} && \heatmap{16.04} & \heatmap{14.33} & \heatmap{15.90} && \heatmap{10.65} & \heatmap{6.51} & \heatmap{6.32}  \\
SigLIP-ViT-S/14 & \heatmap{0.59} & \heatmap{0.79} & \heatmap{0.69} && \heatmap{0.54} & \heatmap{1.14} & \heatmap{1.51}  && \heatmap{3.56} & \heatmap{3.61} & \heatmap{5.66} && \heatmap{21.99} & \heatmap{26.47} & \heatmap{27.53}   \\
CoCa-ViT-L/14 & \heatmap{0.68} & \heatmap{2.09} & \heatmap{2.81} && \heatmap{1.16} & \heatmap{4.28} & \heatmap{7.08} && \heatmap{16.46} & \heatmap{17.34} & \heatmap{16.57} && \heatmap{29.75} & \heatmap{31.22} & \heatmap{31.02}  \\
\bottomrule
\end{tabularx}
\vspace{-5pt}
\label{tab:clip-consistency-main}
\end{table*}

\textbf{\textit{Observation 1.3.} Spurious feature strength partly explains bias shifts, while model factors matter. }
\Cref{fig:acc-delta-corr-conv} (left) shows the correlation between spurious feature strength (\ie, $\text{Acc}_b$ in \Cref{tab:conf-detection-conv}) and $\Delta$. The positive correlation ($r = 0.31$) supports our framework: features more strongly correlated with gender tend to induce larger shifts in bias measurements when perturbed. However, the moderate correlation suggests that model architecture also plays a key role. For instance, while LLaVA-OneVision-7B is highly sensitive to background perturbations on MIAP, other models are more resilient. These findings highlight that effective bias evaluation must account for both spurious features and model-specific vulnerabilities to these features.

\subsection{Experiments: CLIP Variants}

\begin{figure*}[t]
  \centering
  \includegraphics[clip, width=0.86\textwidth]{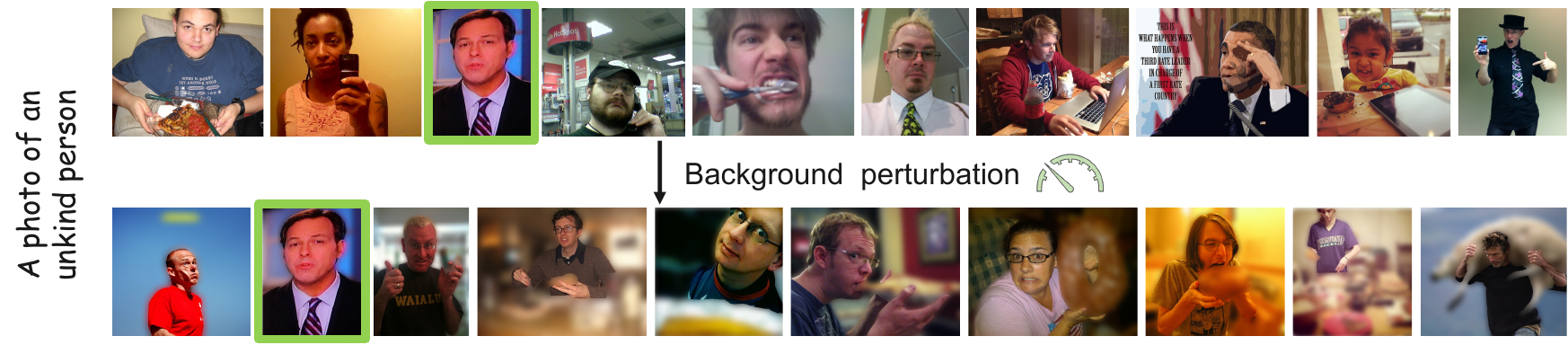}
  \vspace{-8pt}
  \caption{Top-$10$ retrieved images by SigLIP-ViT-S/14 for the prompt ``A photo of an unkind person'' on original and background blurred images (weak perturbation). Green-bordered pairs indicate images retrieved in both sets. The minimal overlap (only one shared image) highlights the model's sensitivity to background changes.  }
  \label{fig:clip-consistency}
  \vspace{-8pt}
\end{figure*}

We also analyze CLIP variants, evaluating how feature perturbations affect gender bias measurements using the MaxSkew metric in \Cref{sec:bias_eval}. 

\vspace{3pt}
\noindent
\textbf{Experimental settings. }
We benchmark five CLIP variants: the original CLIP~\cite{radford2021learning} with ViT-B/32, ViT-L/14, and ViT-H/14 backbones, SigLIP-ViT-S/14 \cite{zhai2023sigmoid}, and CoCa-ViT-L/14 \cite{yu2022coca}. For MaxSkew, we create gender-balanced test sets by sampling equal numbers of male and female images across $5$ random seeds and report averages. Following prior work \cite{seth2023dear,hirota2024saner}, we evaluate using gender-neutral prompts in two categories: adjectives (\eg, ``A photo of an unkind person'') and occupations (\eg, ``A photo of a doctor''). We compute MaxSkew@$1000$ on original and perturbed images to measure $\Delta$, which further details Appendix C.3.

\vspace{3pt}
\noindent
\textbf{Results. }
\Cref{tab:clip-consistency-main} shows the relative difference between MaxSkew@$1000$ for the original and perturbed images. Further details, including the original MaxSkew values for each model and benchmark, are in Appendix D.2. We summarize the main findings below.

\textbf{\textit{Observation 2.1.} CLIP variants exhibit similar vulnerabilities to generative VLMs. }
Like generative VLMs, CLIP variants are most sensitive to object and background perturbations, with object modifications shifting bias up to $25.03\%$ for ViT-B/32 on FACET and background changes causing up to $51.01\%$ for ViT-B/32 on PHASE. \Cref{fig:clip-consistency} further demonstrates this, where weak background blurring drastically alters the retrieval results, confirming the strong dependence on the background. While color and lighting perturbations generally have less impact, CLIP models show distinct sensitivity patterns: color perturbations typically induce moderate shifts ($<5\%$), while lighting effects remain minimal ($<2\%$). These findings highlight that the $G \rightarrow B \rightarrow O$ path distorts bias evaluation across different types of VLMs, making these benchmarks unreliable for measuring gender bias.

\begin{figure}[t]
  \centering
  \includegraphics[clip, width=\columnwidth]{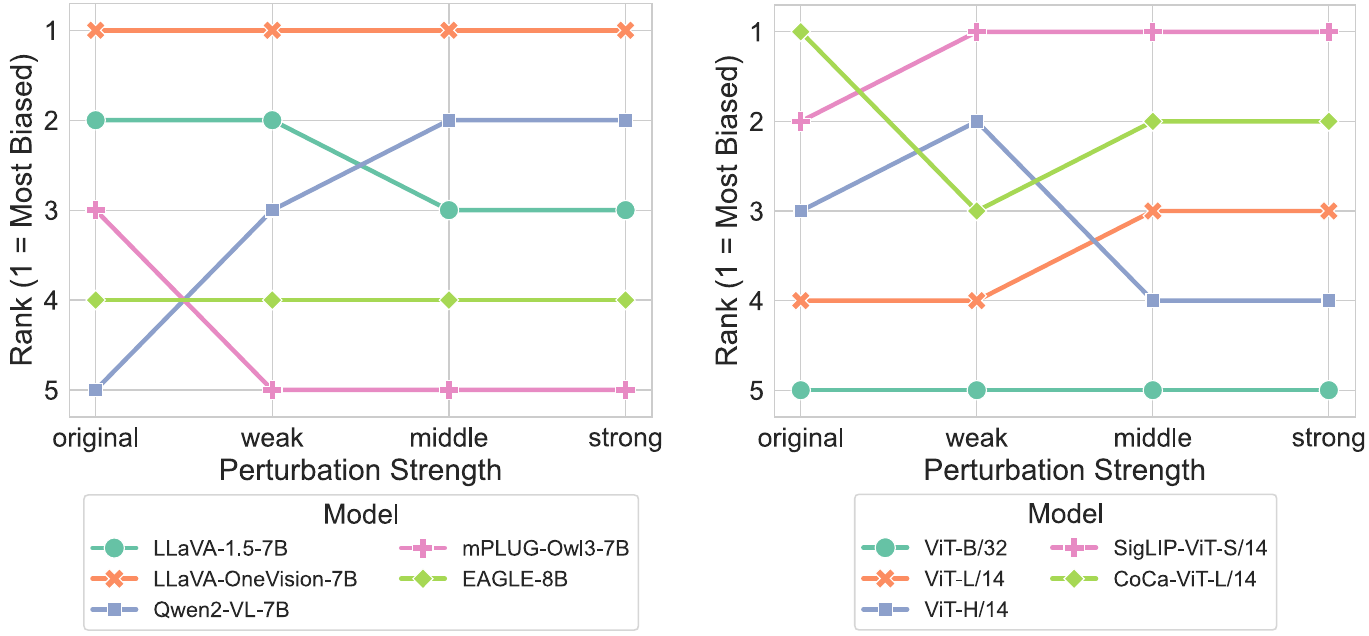}
  \vspace{-19pt}
  \caption{Relative bias ranking changes of generative VLMs under object perturbations on FACET (left) and CLIP variants under background perturbations on COCO-gender (right), based on YGap and MaxSkew values, respectively. Complete results are in Appendix D.2. Feature perturbations alter rankings, raising concerns about the reliability of current evaluations.}
  \label{fig:ranking-main}
  \vspace{-9pt}
\end{figure}

\textbf{\textit{Observation 2.2.} CLIP variants show a stronger correlation between spurious feature strength and bias impact than generative VLMs. }
As shown in \Cref{fig:acc-delta-corr-conv} (right), the correlation between spurious feature strength and bias impact is higher for CLIP models (Pearson's $r = 0.54$) than for generative VLMs ($r = 0.31$). This suggests that CLIP models may rely more on spurious features, making their bias measurements more susceptible to dataset artifacts. These findings further confirm the $G \rightarrow B \rightarrow O$ path as a fundamental challenge in VLM bias evaluation across VLM, highlighting the limitations of current frameworks in reliably measuring actual gender bias.

\textbf{\textit{Observation 2.3.} Feature perturbations drastically alter model rankings in bias benchmarks. }
Beyond sensitivity measurement by $\Delta$, spurious features can completely reshape model comparisons. As shown in \Cref{fig:ranking-main}, perturbing non-gender features can shift the relative ranking of models. For example, in generative VLMs (left), Qwen2-VL moves from the least biased model on original images to the second most biased under object perturbations. This raises concerns about using these benchmarks to identify ``less biased'' models, as rankings may primarily reflect reliance on spurious features rather than actual gender bias. 


\begin{tcolorbox}[colback=white!97!green, colframe=black!70, boxrule=1pt, arc=1mm]
\textbf{Summary}: Spurious features distort gender bias evaluations in VLMs, leading to unreliable bias assessments that reflect the model responses to spurious features rather than true model bias.
\end{tcolorbox}

\section{Toward More Reliable Bias Evaluation}
\label{sec:discussion}

In this section, we examine the fundamental challenges in creating spurious-feature-free benchmarks and propose practical recommendations for more reliable assessment methods that acknowledge these limitations.

\vspace{3pt}
\noindent
\textbf{On the impossibility of creating spurious-feature-free benchmarks. }
Completely eliminating non-gender spurious features from datasets is challenging, as recent studies demonstrate. First, as shown by Wang \etal. \cite{wang2019balanced}, many non-gender features, such as color and background, remain unannotated or are challenging and near-impossible to annotate, making it infeasible to balance the dataset across these features. Second, attempts to aggressively remove gender-correlated features risk eliminating genuinely informative visual cues in a phenomenon known as ``over-adjustment'' \cite{wang2021causal}. Moreover, even synthetic data generation approaches that aim to create counterfactual examples using recent text-to-image generative models can introduce their own biases \cite{mandal2023multimodal,zhang2023auditing,wang2023t2iat,struppek2022biased,ungless2023stereotypes}. As we demonstrated in \Cref{sec:consistency}, VLMs inevitably exploit spurious features in their predictions, rendering bias measurements on the existing benchmarks unreliable indicators of actual gender bias.

\vspace{3pt}
\noindent
\textbf{Recommendations toward more reliable gender bias evaluation. }
Given the demonstrated difficulty of creating spurious-feature-free benchmarks, \textbf{we recommend developing evaluation protocols that explicitly account for the presence of spurious features} rather than assuming they can be eliminated. Specifically, we suggest reporting bias metrics alongside feature-sensitivity measurements derived from controlled perturbation tests like those in our study (\ie, $\Delta$ values). 
One practical approach is to create a two-dimensional evaluation framework where models are assessed on both their measured bias scores (\eg, YGap, MaxSkew) and their stability under perturbations (average $\Delta$ across feature types). For instance, a model with YGap $= 0.05$ and average $\Delta = 10\%$ would be considered more reliably unbiased than one with YGap $= 0.03$ but $\Delta = 50\%$. Additionally, we could compute a composite score $\beta = \text{Bias} \times (1 + \alpha \Delta)$, where $\alpha$ is a weighting parameter that prioritizes stability. 
 This approach enables users to quantify the reliability of bias evaluations themselves, distinguishing between genuinely less-biased models and those that merely appear unbiased because of specific non-gender feature conditions.

\vspace{-3pt}

\begin{tcolorbox}[colback=white!97!green, colframe=black!70, boxrule=1pt, arc=1mm]
\textbf{Summary}: Creating spurious-feature-free benchmarks is fundamentally challenging, requiring new evaluation methods that explicitly account for how non-gender features affect bias measurements.
\end{tcolorbox}

\vspace{-7pt}
\section{Conclusion}
We examine how non-gender spurious features in benchmarks distort gender bias measurements in VLMs. Our controlled perturbation experiments show that even minor modifications to color, objects, and backgrounds can greatly affect bias metrics, with stronger gender-correlated features causing larger distortions. These results confirm that current gender bias evaluations reflect responses to spurious features rather than actual gender bias, undermining their reliability. To address this, we recommend reporting gender bias metrics alongside feature-sensitivity measurements to enable more reliable assessments in the presence of spurious features.\footnote{Appendix F provides further discussion, including limitations.}

{
    \small
    \bibliographystyle{ieeenat_fullname}
    \bibliography{main}

\begin{thebibliography}{75}
\providecommand{\natexlab}[1]{#1}
\providecommand{\url}[1]{\texttt{#1}}
\expandafter\ifx\csname urlstyle\endcsname\relax
  \providecommand{\doi}[1]{doi: #1}\else
  \providecommand{\doi}{doi: \begingroup \urlstyle{rm}\Url}\fi

\bibitem[Bansal et~al.(2022)Bansal, Yin, Monajatipoor, and Chang]{bansal2022well}
Hritik Bansal, Da Yin, Masoud Monajatipoor, and Kai-Wei Chang.
\newblock How well can text-to-image generative models understand ethical natural language interventions?
\newblock In \emph{EMNLP}, 2022.

\bibitem[Berg et~al.(2022)Berg, Hall, Bhalgat, Yang, Kirk, Shtedritski, and Bain]{berg2022prompt}
Hugo Berg, Siobhan~Mackenzie Hall, Yash Bhalgat, Wonsuk Yang, Hannah~Rose Kirk, Aleksandar Shtedritski, and Max Bain.
\newblock A prompt array keeps the bias away: Debiasing vision-language models with adversarial learning.
\newblock In \emph{AACL}, 2022.

\bibitem[Bianchi et~al.(2023)Bianchi, Kalluri, Durmus, Ladhak, Cheng, Nozza, Hashimoto, Jurafsky, Zou, and Caliskan]{bianchi2023easily}
Federico Bianchi, Pratyusha Kalluri, Esin Durmus, Faisal Ladhak, Myra Cheng, Debora Nozza, Tatsunori Hashimoto, Dan Jurafsky, James Zou, and Aylin Caliskan.
\newblock Easily accessible text-to-image generation amplifies demographic stereotypes at large scale.
\newblock In \emph{FAccT}, 2023.

\bibitem[Birhane et~al.(2021)Birhane, Prabhu, and Kahembwe]{birhane2021multimodal}
Abeba Birhane, Vinay~Uday Prabhu, and Emmanuel Kahembwe.
\newblock Multimodal datasets: Misogyny, pornography, and malignant stereotypes.
\newblock \emph{arXiv preprint arXiv:2110.01963}, 2021.

\bibitem[Chen et~al.(2024)Chen, Wang, Cao, Liu, Gao, Cui, Zhu, Ye, Tian, Liu, et~al.]{internvl2}
Zhe Chen, Weiyun Wang, Yue Cao, Yangzhou Liu, Zhangwei Gao, Erfei Cui, Jinguo Zhu, Shenglong Ye, Hao Tian, Zhaoyang Liu, et~al.
\newblock Expanding performance boundaries of open-source multimodal models with model, data, and test-time scaling.
\newblock \emph{arXiv preprint arXiv:2412.05271}, 2024.

\bibitem[Cho et~al.(2023)Cho, Zala, and Bansal]{cho2023dall}
Jaemin Cho, Abhay Zala, and Mohit Bansal.
\newblock Dall-eval: Probing the reasoning skills and social biases of text-to-image generation models.
\newblock In \emph{ICCV}, 2023.

\bibitem[Chuang et~al.(2023)Chuang, Jampani, Li, Torralba, and Jegelka]{chuang2023debiasing}
Ching-Yao Chuang, Varun Jampani, Yuanzhen Li, Antonio Torralba, and Stefanie Jegelka.
\newblock Debiasing vision-language models via biased prompts.
\newblock \emph{arXiv preprint arXiv:2302.00070}, 2023.

\bibitem[Dehdashtian et~al.(2024)Dehdashtian, Wang, and Boddeti]{dehdashtian2024fairvlm}
Sepehr Dehdashtian, Lan Wang, and Vishnu Boddeti.
\newblock Fair{VLM}: Mitigating bias in pre-trained vision-language models.
\newblock In \emph{ICLR}, 2024.

\bibitem[Dehouche(2021)]{dehouche2021implicit}
Nassim Dehouche.
\newblock Implicit stereotypes in pre-trained classifiers.
\newblock \emph{IEEE Access}, 2021.

\bibitem[Fraser and Kiritchenko(2024)]{fraser2024examining}
Kathleen~C Fraser and Svetlana Kiritchenko.
\newblock Examining gender and racial bias in large vision-language models using a novel dataset of parallel images.
\newblock In \emph{EACL}, 2024.

\bibitem[Friedrich et~al.(2023)Friedrich, Schramowski, Brack, Struppek, Hintersdorf, Luccioni, and Kersting]{friedrich2023fair}
Felix Friedrich, Patrick Schramowski, Manuel Brack, Lukas Struppek, Dominik Hintersdorf, Sasha Luccioni, and Kristian Kersting.
\newblock Fair diffusion: Instructing text-to-image generation models on fairness.
\newblock \emph{arXiv preprint arXiv:2302.10893}, 2023.

\bibitem[Garcia et~al.(2023)Garcia, Hirota, Wu, and Nakashima]{garcia2023uncurated}
Noa Garcia, Yusuke Hirota, Yankun Wu, and Yuta Nakashima.
\newblock Uncurated image-text datasets: Shedding light on demographic bias.
\newblock In \emph{CVPR}, 2023.

\bibitem[Geyik et~al.(2019)Geyik, Ambler, and Kenthapadi]{geyik2019fairness}
Sahin~Cem Geyik, Stuart Ambler, and Krishnaram Kenthapadi.
\newblock Fairness-aware ranking in search \& recommendation systems with application to linkedin talent search.
\newblock In \emph{SIGKDD}, 2019.

\bibitem[Girrbach et~al.(2025)Girrbach, Huang, Alaniz, Darrell, and Akata]{girrbach2024revealing}
Leander Girrbach, Yiran Huang, Stephan Alaniz, Trevor Darrell, and Zeynep Akata.
\newblock Revealing and reducing gender biases in vision and language assistants (vlas).
\newblock In \emph{ICLR}, 2025.

\bibitem[Gustafson et~al.(2023)Gustafson, Rolland, Ravi, Duval, Adcock, Fu, Hall, and Ross]{gustafson2023facet}
Laura Gustafson, Chloe Rolland, Nikhila Ravi, Quentin Duval, Aaron Adcock, Cheng-Yang Fu, Melissa Hall, and Candace Ross.
\newblock Facet: Fairness in computer vision evaluation benchmark.
\newblock In \emph{ICCV}, 2023.

\bibitem[Hall et~al.(2023)Hall, Gustafson, Adcock, Misra, and Ross]{hall2023vision}
Melissa Hall, Laura Gustafson, Aaron Adcock, Ishan Misra, and Candace Ross.
\newblock Vision-language models performing zero-shot tasks exhibit gender-based disparities.
\newblock In \emph{ICCV Workshops}, 2023.

\bibitem[Hausladen et~al.(2025)Hausladen, Knott, Camerer, and Perona]{hausladen2025social}
Carina~I Hausladen, Manuel Knott, Colin~F Camerer, and Pietro Perona.
\newblock Social perception of faces in a vision-language model.
\newblock In \emph{FAccT}, 2025.

\bibitem[He et~al.(2016)He, Zhang, Ren, and Sun]{resnet}
Kaiming He, Xiangyu Zhang, Shaoqing Ren, and Jian Sun.
\newblock Identity mappings in deep residual networks.
\newblock In \emph{ECCV}, 2016.

\bibitem[Hirota et~al.(2022)Hirota, Nakashima, and Garcia]{hirota2022quantifying}
Yusuke Hirota, Yuta Nakashima, and Noa Garcia.
\newblock Quantifying societal bias amplification in image captioning.
\newblock In \emph{CVPR}, 2022.

\bibitem[Hirota et~al.(2024{\natexlab{a}})Hirota, Andrews, Zhao, Papakyriakopoulos, Modas, Nakashima, and Xiang]{hirota2024resampled}
Yusuke Hirota, Jerone~TA Andrews, Dora Zhao, Orestis Papakyriakopoulos, Apostolos Modas, Yuta Nakashima, and Alice Xiang.
\newblock Resampled datasets are not enough: Mitigating societal bias beyond single attributes.
\newblock In \emph{EMNLP}, 2024{\natexlab{a}}.

\bibitem[Hirota et~al.(2024{\natexlab{b}})Hirota, Hachiuma, Yang, and Nakashima]{hirota2024descriptive}
Yusuke Hirota, Ryo Hachiuma, Chao-Han~Huck Yang, and Yuta Nakashima.
\newblock From descriptive richness to bias: Unveiling the dark side of generative image caption enrichment.
\newblock In \emph{EMNLP}, 2024{\natexlab{b}}.

\bibitem[Hirota et~al.(2025)Hirota, Chen, Wang, Nakashima, Wang, and Hachiuma]{hirota2024saner}
Yusuke Hirota, Min-Hung Chen, Chien-Yi Wang, Yuta Nakashima, Yu-Chiang~Frank Wang, and Ryo Hachiuma.
\newblock Saner: Annotation-free societal attribute neutralizer for debiasing clip.
\newblock In \emph{ICLR}, 2025.

\bibitem[Howard et~al.(2024)Howard, Madasu, Le, Moreno, Bhiwandiwalla, and Lal]{howard2023probing}
Phillip Howard, Avinash Madasu, Tiep Le, Gustavo~Lujan Moreno, Anahita Bhiwandiwalla, and Vasudev Lal.
\newblock Probing and mitigating intersectional social biases in vision-language models with counterfactual examples.
\newblock In \emph{CVPR}, 2024.

\bibitem[Howard et~al.(2025)Howard, Bhiwandiwalla, Fraser, and Kiritchenko]{howard2024uncovering}
Phillip Howard, Anahita Bhiwandiwalla, Kathleen~C Fraser, and Svetlana Kiritchenko.
\newblock Uncovering bias in large vision-language models with counterfactuals.
\newblock In \emph{NAACL}, 2025.

\bibitem[Jiang et~al.(2024)Jiang, Li, Shen, Liu, Backes, and Zhang]{jiang2024texttt}
Yukun Jiang, Zheng Li, Xinyue Shen, Yugeng Liu, Michael Backes, and Yang Zhang.
\newblock Modscan: Measuring stereotypical bias in large vision-language models from vision and language modalities.
\newblock In \emph{EMNLP}, 2024.

\bibitem[Krasin et~al.(2017)Krasin, Duerig, Alldrin, Ferrari, Abu-El-Haija, Kuznetsova, Rom, Uijlings, Popov, Veit, et~al.]{krasin2017openimages}
Ivan Krasin, Tom Duerig, Neil Alldrin, Vittorio Ferrari, Sami Abu-El-Haija, Alina Kuznetsova, Hassan Rom, Jasper Uijlings, Stefan Popov, Andreas Veit, et~al.
\newblock Openimages: A public dataset for large-scale multi-label and multi-class image classification.
\newblock \emph{Dataset available from https://github. com/openimages}, 2017.

\bibitem[Li et~al.(2024)Li, Zhang, Guo, Zhang, Li, Zhang, Zhang, Zhang, Li, Liu, et~al.]{llavaonevison}
Bo Li, Yuanhan Zhang, Dong Guo, Renrui Zhang, Feng Li, Hao Zhang, Kaichen Zhang, Peiyuan Zhang, Yanwei Li, Ziwei Liu, et~al.
\newblock Llava-onevision: Easy visual task transfer.
\newblock \emph{TMLR}, 2024.

\bibitem[Lin et~al.(2014)Lin, Maire, Belongie, Hays, Perona, Ramanan, Doll{\'a}r, and Zitnick]{lin2014microsoft}
Tsung-Yi Lin, Michael Maire, Serge Belongie, James Hays, Pietro Perona, Deva Ramanan, Piotr Doll{\'a}r, and C~Lawrence Zitnick.
\newblock Microsoft {COCO}: Common objects in context.
\newblock In \emph{ECCV}, 2014.

\bibitem[Liu et~al.(2024{\natexlab{a}})Liu, Li, Wu, and Lee]{liu2024visual}
Haotian Liu, Chunyuan Li, Qingyang Wu, and Yong~Jae Lee.
\newblock Visual instruction tuning.
\newblock \emph{NeurIPS}, 2024{\natexlab{a}}.

\bibitem[Liu et~al.(2022)Liu, Mao, Wu, Feichtenhofer, Darrell, and Xie]{liu2022convnet}
Zhuang Liu, Hanzi Mao, Chao-Yuan Wu, Christoph Feichtenhofer, Trevor Darrell, and Saining Xie.
\newblock A convnet for the 2020s.
\newblock In \emph{CVPR}, 2022.

\bibitem[Liu et~al.(2024{\natexlab{b}})Liu, Schaldenbrand, Okogwu, Peng, Yun, Hundt, Kim, and Oh]{liu2024scoft}
Zhixuan Liu, Peter Schaldenbrand, Beverley-Claire Okogwu, Wenxuan Peng, Youngsik Yun, Andrew Hundt, Jihie Kim, and Jean Oh.
\newblock Scoft: Self-contrastive fine-tuning for equitable image generation.
\newblock In \emph{CVPR}, 2024{\natexlab{b}}.

\bibitem[Luccioni et~al.(2023)Luccioni, Akiki, Mitchell, and Jernite]{luccioni2023stable}
Alexandra~Sasha Luccioni, Christopher Akiki, Margaret Mitchell, and Yacine Jernite.
\newblock Stable bias: Analyzing societal representations in diffusion models.
\newblock In \emph{NeurIPS}, 2023.

\bibitem[Mandal et~al.(2023)Mandal, Leavy, and Little]{mandal2023multimodal}
Abhishek Mandal, Susan Leavy, and Suzanne Little.
\newblock Multimodal composite association score: Measuring gender bias in generative multimodal models.
\newblock \emph{arXiv preprint arXiv:2304.13855}, 2023.

\bibitem[Meister et~al.(2023)Meister, Zhao, Wang, Ramaswamy, Fong, and Russakovsky]{meister2023gender}
Nicole Meister, Dora Zhao, Angelina Wang, Vikram~V Ramaswamy, Ruth Fong, and Olga Russakovsky.
\newblock Gender artifacts in visual datasets.
\newblock In \emph{ICCV}, 2023.

\bibitem[Naik and Nushi(2023)]{naik2023social}
Ranjita Naik and Besmira Nushi.
\newblock Social biases through the text-to-image generation lens.
\newblock In \emph{AIES}, 2023.

\bibitem[Nair and Hinton(2010)]{nair2010rectified}
Vinod Nair and Geoffrey~E Hinton.
\newblock Rectified linear units improve restricted boltzmann machines.
\newblock In \emph{ICML}, 2010.

\bibitem[OpenAI(2023)]{2023GPT4VisionSC}
OpenAI.
\newblock Gpt-4v(ision) system card, 2023.

\bibitem[Qiu et~al.(2023)Qiu, Dou, Wang, Celikyilmaz, and Peng]{qiu2023gender}
Haoyi Qiu, Zi-Yi Dou, Tianlu Wang, Asli Celikyilmaz, and Nanyun Peng.
\newblock Gender biases in automatic evaluation metrics for image captioning.
\newblock In \emph{EMNLP}, 2023.

\bibitem[Radford et~al.(2021)Radford, Kim, Hallacy, Ramesh, Goh, Agarwal, Sastry, Askell, Mishkin, Clark, et~al.]{radford2021learning}
Alec Radford, Jong~Wook Kim, Chris Hallacy, Aditya Ramesh, Gabriel Goh, Sandhini Agarwal, Girish Sastry, Amanda Askell, Pamela Mishkin, Jack Clark, et~al.
\newblock Learning transferable visual models from natural language supervision.
\newblock In \emph{ICML}, 2021.

\bibitem[Raj et~al.(2024)Raj, Mukherjee, Caliskan, Anastasopoulos, and Zhu]{raj2024biasdora}
Chahat Raj, Anjishnu Mukherjee, Aylin Caliskan, Antonios Anastasopoulos, and Ziwei Zhu.
\newblock Biasdora: Exploring hidden biased associations in vision-language models.
\newblock In \emph{EMNLP findings}, 2024.

\bibitem[Rombach et~al.(2022)Rombach, Blattmann, Lorenz, Esser, and Ommer]{rombach2022high}
Robin Rombach, Andreas Blattmann, Dominik Lorenz, Patrick Esser, and Bj{\"o}rn Ommer.
\newblock High-resolution image synthesis with latent diffusion models.
\newblock In \emph{CVPR}, 2022.

\bibitem[Ross et~al.(2021)Ross, Katz, and Barbu]{ross2020measuring}
Candace Ross, Boris Katz, and Andrei Barbu.
\newblock Measuring social biases in grounded vision and language embeddings.
\newblock In \emph{ACL}, 2021.

\bibitem[Ruggeri et~al.(2023)Ruggeri, Nozza, et~al.]{ruggeri2023multi}
Gabriele Ruggeri, Debora Nozza, et~al.
\newblock A multi-dimensional study on bias in vision-language models.
\newblock In \emph{Findings of ACL}, 2023.

\bibitem[Sathe et~al.(2024)Sathe, Jain, and Sitaram]{sathe2024unified}
Ashutosh Sathe, Prachi Jain, and Sunayana Sitaram.
\newblock A unified framework and dataset for assessing societal bias in vision-language models.
\newblock In \emph{EMNLP findings}, 2024.

\bibitem[Schumann et~al.(2021)Schumann, Ricco, Prabhu, Ferrari, and Pantofaru]{schumann2021step}
Candice Schumann, Susanna Ricco, Utsav Prabhu, Vittorio Ferrari, and Caroline Pantofaru.
\newblock A step toward more inclusive people annotations for fairness.
\newblock In \emph{AIES}, 2021.

\bibitem[Seshadri et~al.(2023)Seshadri, Singh, and Elazar]{seshadri2023bias}
Preethi Seshadri, Sameer Singh, and Yanai Elazar.
\newblock The bias amplification paradox in text-to-image generation.
\newblock \emph{arXiv preprint arXiv:2308.00755}, 2023.

\bibitem[Seth et~al.(2023)Seth, Hemani, and Agarwal]{seth2023dear}
Ashish Seth, Mayur Hemani, and Chirag Agarwal.
\newblock Dear: Debiasing vision-language models with additive residuals.
\newblock In \emph{CVPR}, 2023.

\bibitem[Sharma et~al.(2018)Sharma, Ding, Goodman, and Soricut]{sharma2018conceptual}
Piyush Sharma, Nan Ding, Sebastian Goodman, and Radu Soricut.
\newblock Conceptual captions: A cleaned, hypernymed, image alt-text dataset for automatic image captioning.
\newblock In \emph{ACL}, 2018.

\bibitem[Shi et~al.(2025)Shi, Liu, Wang, Liao, Radhakrishnan, Huang, Yin, Sapra, Yacoob, Shi, et~al.]{shi2024eagle}
Min Shi, Fuxiao Liu, Shihao Wang, Shijia Liao, Subhashree Radhakrishnan, De-An Huang, Hongxu Yin, Karan Sapra, Yaser Yacoob, Humphrey Shi, et~al.
\newblock Eagle: Exploring the design space for multimodal llms with mixture of encoders.
\newblock In \emph{ICLR}, 2025.

\bibitem[Slyman et~al.(2024)Slyman, Lee, Cohen, and Kafle]{slyman2024fairdedup}
Eric Slyman, Stefan Lee, Scott Cohen, and Kushal Kafle.
\newblock Fairdedup: Detecting and mitigating vision-language fairness disparities in semantic dataset deduplication.
\newblock In \emph{CVPR}, 2024.

\bibitem[Srinivasan and Bisk(2022)]{srinivasan2021worst}
Tejas Srinivasan and Yonatan Bisk.
\newblock Worst of both worlds: Biases compound in pre-trained vision-and-language models.
\newblock In \emph{ACL Workshops}, 2022.

\bibitem[Struppek et~al.(2022)Struppek, Hintersdorf, and Kersting]{struppek2022biased}
Lukas Struppek, Dominik Hintersdorf, and Kristian Kersting.
\newblock The biased artist: Exploiting cultural biases via homoglyphs in text-guided image generation models.
\newblock \emph{arXiv preprint arXiv:2209.08891}, 2022.

\bibitem[Tanjim et~al.(2024)Tanjim, Singh, Kafle, Sinha, and Cottrell]{tanjim2024discovering}
Md~Mehrab Tanjim, Krishna~Kumar Singh, Kushal Kafle, Ritwik Sinha, and Garrison~W Cottrell.
\newblock Discovering and mitigating biases in clip-based image editing.
\newblock In \emph{WACV}, 2024.

\bibitem[Ungless et~al.(2023)Ungless, Ross, and Lauscher]{ungless2023stereotypes}
Eddie~L Ungless, Bj{\"o}rn Ross, and Anne Lauscher.
\newblock Stereotypes and smut: The (mis) representation of non-cisgender identities by text-to-image models.
\newblock In \emph{ACL}, 2023.

\bibitem[Wang and Russakovsky(2021)]{wang2021biasamp}
Angelina Wang and Olga Russakovsky.
\newblock Directional bias amplification.
\newblock In \emph{ICML}, 2021.

\bibitem[Wang and Russakovsky(2023)]{wang2023overwriting}
Angelina Wang and Olga Russakovsky.
\newblock Overwriting pretrained bias with finetuning data.
\newblock In \emph{ICCV}, 2023.

\bibitem[Wang et~al.(2021{\natexlab{a}})Wang, Liu, and Wang]{wang2021gender}
Jialu Wang, Yang Liu, and Xin~Eric Wang.
\newblock Are gender-neutral queries really gender-neutral? mitigating gender bias in image search.
\newblock \emph{arXiv preprint arXiv:2109.05433}, 2021{\natexlab{a}}.

\bibitem[Wang et~al.(2023{\natexlab{a}})Wang, Liu, Di, Liu, and Wang]{wang2023t2iat}
Jialu Wang, Xinyue~Gabby Liu, Zonglin Di, Yang Liu, and Xin~Eric Wang.
\newblock T2iat: Measuring valence and stereotypical biases in text-to-image generation.
\newblock In \emph{ACL}, 2023{\natexlab{a}}.

\bibitem[Wang et~al.(2024{\natexlab{a}})Wang, Bai, Tan, Wang, Fan, Bai, Chen, Liu, Wang, Ge, et~al.]{wang2024qwen2}
Peng Wang, Shuai Bai, Sinan Tan, Shijie Wang, Zhihao Fan, Jinze Bai, Keqin Chen, Xuejing Liu, Jialin Wang, Wenbin Ge, et~al.
\newblock Qwen2-vl: Enhancing vision-language model's perception of the world at any resolution.
\newblock \emph{arXiv preprint arXiv:2409.12191}, 2024{\natexlab{a}}.

\bibitem[Wang et~al.(2024{\natexlab{b}})Wang, Cao, Zhang, Yuan, Shan, Chen, and Gao]{wang2024vlbiasbench}
Sibo Wang, Xiangkui Cao, Jie Zhang, Zheng Yuan, Shiguang Shan, Xilin Chen, and Wen Gao.
\newblock Vlbiasbench: A comprehensive benchmark for evaluating bias in large vision-language model.
\newblock \emph{arXiv preprint arXiv:2406.14194}, 2024{\natexlab{b}}.

\bibitem[Wang et~al.(2019)Wang, Zhao, Yatskar, Chang, and Ordonez]{wang2019balanced}
Tianlu Wang, Jieyu Zhao, Mark Yatskar, Kai-Wei Chang, and Vicente Ordonez.
\newblock Balanced datasets are not enough: Estimating and mitigating gender bias in deep image representations.
\newblock In \emph{ICCV}, 2019.

\bibitem[Wang et~al.(2021{\natexlab{b}})Wang, Zhou, Sun, and Zhang]{wang2021causal}
Tan Wang, Chang Zhou, Qianru Sun, and Hanwang Zhang.
\newblock Causal attention for unbiased visual recognition.
\newblock In \emph{CVPR}, 2021{\natexlab{b}}.

\bibitem[Wang et~al.(2023{\natexlab{b}})Wang, Yi, Jiang, Zhou, Wei, and Xie]{wang2023tovilag}
Xinpeng Wang, Xiaoyuan Yi, Han Jiang, Shanlin Zhou, Zhihua Wei, and Xing Xie.
\newblock Tovilag: Your visual-language generative model is also an evildoer.
\newblock In \emph{EMNLP}, 2023{\natexlab{b}}.

\bibitem[Weng et~al.(2024)Weng, Gao, Andrews, and Zhao]{weng2024images}
Zhaotian Weng, Zijun Gao, Jerone Andrews, and Jieyu Zhao.
\newblock Images speak louder than words: Understanding and mitigating bias in vision-language model from a causal mediation perspective.
\newblock In \emph{EMNLP}, 2024.

\bibitem[Wolfe et~al.(2023)Wolfe, Yang, Howe, and Caliskan]{wolfe2023contrastive}
Robert Wolfe, Yiwei Yang, Bill Howe, and Aylin Caliskan.
\newblock Contrastive language-vision ai models pretrained on web-scraped multimodal data exhibit sexual objectification bias.
\newblock In \emph{FAccT}, 2023.

\bibitem[Xiao et~al.(2024)Xiao, Liu, Cheng, Yin, Liang, Li, Shao, Liu, and Tao]{xiao2024genderbias}
Yisong Xiao, Aishan Liu, QianJia Cheng, Zhenfei Yin, Siyuan Liang, Jiapeng Li, Jing Shao, Xianglong Liu, and Dacheng Tao.
\newblock Genderbias-$\backslash$emph $\{$VL$\}$: Benchmarking gender bias in vision language models via counterfactual probing.
\newblock \emph{arXiv preprint arXiv:2407.00600}, 2024.

\bibitem[Ye et~al.(2025)Ye, Xu, Liu, Hu, Yan, Qian, Zhang, Huang, and Zhou]{ye2024mplug}
Jiabo Ye, Haiyang Xu, Haowei Liu, Anwen Hu, Ming Yan, Qi Qian, Ji Zhang, Fei Huang, and Jingren Zhou.
\newblock mplug-owl3: Towards long image-sequence understanding in multi-modal large language models.
\newblock In \emph{ICLR}, 2025.

\bibitem[Yu et~al.(2022)Yu, Wang, Vasudevan, Yeung, Seyedhosseini, and Wu]{yu2022coca}
Jiahui Yu, Zirui Wang, Vijay Vasudevan, Legg Yeung, Mojtaba Seyedhosseini, and Yonghui Wu.
\newblock Coca: Contrastive captioners are image-text foundation models.
\newblock \emph{TMLR}, 2022.

\bibitem[Zeng et~al.(2025)Zeng, Yin, and Liu]{zeng2025understanding}
Boya Zeng, Yida Yin, and Zhuang Liu.
\newblock Understanding bias in large-scale visual datasets.
\newblock \emph{NeurIPS}, 2025.

\bibitem[Zhai et~al.(2023)Zhai, Mustafa, Kolesnikov, and Beyer]{zhai2023sigmoid}
Xiaohua Zhai, Basil Mustafa, Alexander Kolesnikov, and Lucas Beyer.
\newblock Sigmoid loss for language image pre-training.
\newblock In \emph{ICCV}, 2023.

\bibitem[Zhang et~al.(2025)Zhang, Guo, and Kankanhalli]{zhang2024joint}
Haoyu Zhang, Yangyang Guo, and Mohan Kankanhalli.
\newblock Joint vision-language social bias removal for clip.
\newblock In \emph{CVPR}, 2025.

\bibitem[Zhang et~al.(2023)Zhang, Jiang, Turk, and Yang]{zhang2023auditing}
Yanzhe Zhang, Lu Jiang, Greg Turk, and Diyi Yang.
\newblock Auditing gender presentation differences in text-to-image models.
\newblock \emph{arXiv preprint arXiv:2302.03675}, 2023.

\bibitem[Zhao et~al.(2021)Zhao, Wang, and Russakovsky]{zhao2021captionbias}
Dora Zhao, Angelina Wang, and Olga Russakovsky.
\newblock Understanding and evaluating racial biases in image captioning.
\newblock In \emph{ICCV}, 2021.

\bibitem[Zhao et~al.(2017)Zhao, Wang, Yatskar, Ordonez, and Chang]{zhao2017mals}
Jieyu Zhao, Tianlu Wang, Mark Yatskar, Vicente Ordonez, and Kai-Wei Chang.
\newblock Men also like shopping: Reducing gender bias amplification using corpus-level constraints.
\newblock In \emph{EMNLP}, 2017.

\bibitem[Zhou et~al.(2022)Zhou, Girdhar, Joulin, Kr{\"a}henb{\"u}hl, and Misra]{zhou2022detecting}
Xingyi Zhou, Rohit Girdhar, Armand Joulin, Philipp Kr{\"a}henb{\"u}hl, and Ishan Misra.
\newblock Detecting twenty-thousand classes using image-level supervision.
\newblock In \emph{ECCV}, 2022.

\end{thebibliography}
}

\clearpage

\appendix

\section*{Appendix}
This appendix includes:

\begin{itemize}
    \item Candidate feature selection (Appendix~\ref{sec:app-feature}).
    \item Further discussion on the perturbation strategy selection (Appendix~\ref{sec:app-perturbation}).
    \item Implementation details (Appendix~\ref{sec:app-details}).
    \item Additional experimental results (Appendix~\ref{sec:app-results}).
    \item Additional visual examples (Appendix~\ref{sec:app-examples}).   
    \item Discussion (Appendix~\ref{sec:app-discussion}).
\end{itemize}

\section{Candidate Feature Selection}
\label{sec:app-feature}

In this study, we focus on four non-gender features—color, lighting, object, and background—as potential spurious features in gender bias evaluation. These are selected based on prior research \cite{meister2023gender,zeng2025understanding} and practical considerations for our perturbation-based methodology.

Meister \etal.~\cite{meister2023gender} identified several visual cues correlated with gender in computer vision datasets. They showed that low-level features, such as color distributions, differ by gender, with images of women often having warmer hues and those of men having cooler tones. Additionally, they found that contextual elements, including objects and backgrounds, can serve as gender predictors even when the person is masked.

We prioritize these four features due to:
\begin{itemize}
    \item \textbf{Established gender correlation}: Prior studies~\cite{meister2023gender,zeng2025understanding} have shown that these features strongly correlate with gender in datasets like COCO and OpenImages. Our results in Section~3 further confirm that color, object, and background enable above-chance gender prediction across all examined benchmarks.
    \item \textbf{Controlled perturbability}: They can be modified in isolation while preserving other image characteristics, allowing precise analysis of their impact.
    \item \textbf{Diversity of feature types}: Our selection covers both low-level properties (color, lighting) and high-level contextual elements (objects, background).
\end{itemize}

While other potential spurious features exist (\eg, image composition, text overlays, photographic style), \textbf{our selection provides a strong foundation for studying spurious features effects while maintaining experimental tractability.}

\section{Selection of Perturbation Strategies}
\label{sec:app-perturbation}

We opt for straightforward, simple image processing-based feature perturbations (\eg, hue shifts, background blurring) rather than text-to-image (T2I) generative model-based feature editing as:
\begin{itemize}
    \item \textbf{T2I models introduce additional biases}. State-of-the-art models like Stable Diffusion~\cite{rombach2022high} are known to encode gender, racial, and cultural stereotypes \cite{naik2023social,seshadri2023bias,friedrich2023fair,liu2024scoft}, raising fairness concerns. \textbf{Using T2I models for perturbations would risk contaminating our study with these biases. }
    \item \textbf{T2I models lack precise control over feature isolation}. While they can generate image variations, maintaining all other aspects constant is difficult. Even with careful prompting, generative models may subtly alter multiple attributes simultaneously, making it challenging to attribute model output changes to specific perturbations. 
\end{itemize} 

In contrast, our selected perturbation techniques ensure controlled and consistent transformations across the dataset. They offer parametric control over perturbation strength (weak, middle, strong) while preserving individual identity and gender-relevant attributes. This allows us to isolate spurious effects with greater precision while avoiding the introduction of additional biases.

\section{Implementation Details}
\label{sec:app-details}

\subsection{Details of Gender Bias Benchmarks}

In this work, we focus on the four well-known gender bias benchmarks: COCO-gender \cite{zhao2021captionbias}, FACET \cite{gustafson2023facet}, MIAP \cite{schumann2021step}, and PHASE \cite{garcia2023uncurated}. These benchmarks provide binary gender annotations for real-world images such as COCO. We detail each benchmark below:
\begin{itemize}
    \item \textbf{COCO-gender} provides human attribute annotations, including gender and skin tone, for the COCO validation set. 
    \item \textbf{FACET} presents more inclusive human attribute annotations, such as gender, age, and hair color, for a subset of Segment Anything 1 Billion.
    \item \textbf{MIAP} presents gender and age annotations for a subset of OpenImages. 
    \item \textbf{PHASE} provides comprehensive attribute annotations (\eg, gender, ethnicity, emotion) for a subset of Google Conceptual Captions. 
\end{itemize}
For all benchmarks, we select images that do not contain multiple people for robust analysis. The statistics of the datasets are shown in \Cref{tab:data-stats}.

\begin{table}[t]
\footnotesize
\centering
\caption{Statistics of gender bias benchmarks used in our experiments. Woman/Man indicates the number of woman/man images. For all datasets, we focus on images that do not contain multiple people for robust analysis.}
\vspace{-5pt}
\begin{tabularx}{0.8\columnwidth}{X r r r r r  }
\toprule
Benchmark & Woman & Man  & Total  \\
\midrule
COCO-gender & 1,568 & 3,156 & 4,724 \\
FACET & 7,009 & 21,776 & 28,785 \\
MIAP & 1,459 & 4,501 & 5,960 \\
PAHSE & 4,031 & 6,831 & 10,862 \\
\bottomrule
\end{tabularx}
\label{tab:data-stats}
\vspace{-2pt}
\end{table}

\subsection{Details of Feature Perturbation}

\textbf{Color perturbation. }
For color perturbation, we alter the Hue component in HSV color space while preserving saturation and value (brightness). We implement different perturbation strengths through controlled random Hue shifts:
\begin{itemize}
    \item Weak perturbation: Random Hue shifts within $\pm10$ units in the HSV Hue channel ($0-255$ scale)
    \item Middle perturbation: Random Hue shifts within $\pm20$ units, excluding the central range of $\pm10$
    \item Strong perturbation: Random Hue shifts within $\pm30$ units, excluding the central range of $\pm10$
\end{itemize}

\vspace{3pt}
\noindent
\textbf{Lighting perturbation. }
For lighting perturbation, we modify the Value (brightness) component in HSV color space while preserving hue and saturation. We implement different perturbation strengths by controlling the range of random Value shifts:
\begin{itemize}
    \item Weak perturbation: Random Value shifts within $\pm10$ units in the HSV Value channel ($0-255$ scale)
    \item Middle perturbation: Random Value shifts within $\pm20$ units, excluding the central range of $\pm10$
    \item Strong perturbation: Random Value shifts within $\pm30$ units, excluding the central range of $\pm10$
\end{itemize}

\vspace{3pt}
\noindent
\textbf{Object perturbation. }
For object perturbation, we randomly mask $10\%$, $20\%$, or $30\%$ objects in images. We do not remove persons to keep gender recognition. The objects are detected by the object detector \cite{zhou2022detecting}.

\vspace{3pt}
\noindent
\textbf{Background perturbation. }
For background perturbation, we selectively apply Gaussian blur to only the background regions of images while preserving the person in focus. We implement different perturbation strengths by varying the blur radius:
\begin{itemize}
    \item Weak perturbation: Gaussian blur with radius$=10$ applied to background areas
    \item Middle perturbation: Gaussian blur with radius$=25$ applied to background areas
    \item Strong perturbation: Gaussian blur with radius$=40$ applied to background areas
\end{itemize}
The implementation uses person masks to ensure that only non-person regions are affected. This approach maintains all gender-relevant visual information about the person while modifying the contextual background, allowing us to isolate the impact of background features on model predictions.

\subsection{Details of Gender Bias Evaluation}

This section provides additional details on the gender bias evaluation methods used in our study, expanding on the information presented in Section 3.1.

\vspace{3pt}
\noindent
\textbf{VQA-based evaluation for generative VLMs. }
Following the methodology in Girrbach \etal. \cite{girrbach2024revealing}, we evaluated generative VLMs using a comprehensive set of questions designed to assess gender bias across multiple domains:
\begin{itemize}
    \item \textbf{Personality traits:} We use $20$ personality traits (\eg, ``friendly'', ``honest'', ``humble'') and a specific question format, ``Is the person in this image [personality trait]''. 
    \item \textbf{Skills:} We included $21$ skill-related questions (\eg, ``work independently'', ``interact with individuals'', ``work in team'') formatted as ``Does this person have the ability to [skill]?''. 
    \item \textbf{Occupations:} We employed $40$ occupation-related questions (\eg, ``laborer'', ``accountant'', ``teacher'') using the format ``Would this person be better suited to be a [occupation]?'' 
\end{itemize}
For each question, an instruction ``A. Yes, B. No, C. Unsure. Indicate the correct option as your answer.'' is added to prompt models to select one of the options. 

For calculating YGap values, we processed model outputs to extract the probability of a ``Yes'' response for each question and image. The YGap was then computed as the difference between the average probability of ``Yes'' responses for men versus women across all images in the dataset.

\vspace{3pt}
\noindent
\textbf{Text-to-image retrieval for CLIP variants. }
Following the previous works \cite{hirota2024saner,berg2022prompt,seth2023dear,zhang2024joint,chuang2023debiasing}, we used a diverse set of gender-neutral prompts across two categories:
\begin{itemize}
    \item \textbf{Adjectives (85 prompts):} Covering various personality traits, both positive (\eg, ``good'', ``kind'', ``smart'') and negative (\eg, "evil", "criminal", "violent"). These were formatted using templates such as ``This is a photo of a [trait] person'' or ``This person is [trait]''. 
    \item \textbf{Occupations (97 prompts):} Spanning diverse professional roles, including both stereotypically male-dominated fields (\eg, ``computer programmer'', ``engineer'', ``CEO'') and stereotypically female-dominated ones (\eg, ``nurse'', ``childcare worker'', ``social worker''). These used templates like ``This is a photo of a [occupation]'' or ``A [occupation]''.
\end{itemize}
For each prompt, we calculated the cosine similarity between the text embedding and all image embeddings in the gender-balanced test set. We then ranked the images based on these similarity scores and computed the MaxSkew@$1000$ metric as described in Section 3.1.
To ensure robustness, we performed each experiment $5$ times with different random seeds for sampling gender-balanced test sets and reported the average MaxSkew values.

\section{Additional Experimental Results}
\label{sec:app-results}

\subsection{Human Study on Gender Recognition Robustness After Perturbations}

\begin{table}[t]
\scriptsize
\centering
\caption{Consistency between human-identified gender and original gender labels for perturbed images. Results show percentage agreement across 200 randomly sampled images from COCO-gender, demonstrating that our feature perturbations preserve gender recognition.}
\vspace{-7pt}
\begin{tabularx}{0.9\columnwidth}{X r r r r r r }
\toprule
Benchmark & Color & Lighting  & Object  & Background \\
\midrule
COCO-gender & 100.0 & 100.0  & 99.6 & 100.0 \\
FACET & 100.0 & 100.0 & 99.8 & 100.0 \\
MIAP & 100.0 & 100.0 & 100.0 & 100.0 \\
PHASE & 100.0 & 100.0 & 99.5 & 100.0 \\
\bottomrule
\end{tabularx}
\label{tab:human-study}
\vspace{-2pt}
\end{table}

To ensure that our feature perturbations do not affect gender recognition, we conducted a human evaluation study. Specifically, we show randomly sampled $200$ feature-perturbed (strong perturbations) images from COCO-gender, asking about the gender of individuals in these images. We then compute the consistency between the answered gender and the gender label of the original images. The results are shown in \Cref{tab:human-study}. While object perturbations occasionally obscure facial features, slightly reducing consistency, overall agreement remains high for all the features.

\subsection{Complete Results of YGap and MaxSkew}


\begin{table*}[t]
\renewcommand{\arraystretch}{1.1}
\setlength{\tabcolsep}{3pt}
\scriptsize
\centering
\caption{YGap results (scaled by $100$) of the generative VLMs. Weak, middle, and strong mean the level of the image perturbation, and original means the results for the original images. Gray cells indicate cases where the original YGap value is nearly $0$ ($\text{YGap}<0.005$ before scaled by $100$), leading to unstable $\Delta$ computation, which we exclude from the analysis in the main paper.
}
\vspace{-5pt}
\setlength{\tabcolsep}{3pt}
\begin{tabularx}{0.95\textwidth}{l r r r r r r r r r r r r r r r r r r r}
\toprule
& \multicolumn{4}{c}{Color} &&\multicolumn{4}{c}{Lighting} &&\multicolumn{4}{c}{Object} &&\multicolumn{4}{c}{Background}
\\ 
\cline{2-5} 
\cline{7-10}
\cline{12-15}
\cline{17-20}
\multirow{-0.8}{*}{Model} & \multirow{1.3}{*}{original} & \multirow{1.3}{*}{weak} & \multirow{1.3}{*}{middle} & \multirow{1.3}{*}{strong} && \multirow{1.3}{*}{original} & \multirow{1.3}{*}{weak} & \multirow{1.3}{*}{middle} & \multirow{1.3}{*}{strong} && \multirow{1.3}{*}{original} & \multirow{1.3}{*}{weak} & \multirow{1.3}{*}{middle} & \multirow{1.3}{*}{strong} && \multirow{1.3}{*}{original} & \multirow{1.3}{*}{weak} & \multirow{1.3}{*}{middle} & \multirow{1.3}{*}{strong}\\
\midrule
\textbf{\textit{COCO-gender}} &  &  &  &  &&  &   &  &  &&  &  &  &  &&  &  &  &   \\
LLaVA-1.5-7B & -3.69  & -3.63  & -3.58 & -3.47  && -3.69 & -3.57  & -3.65 & -3.62 && -3.69 & -3.22 & -2.70 & -2.52 && -3.69  & -3.88  & -4.15 & -4.05   \\
LLaVA-OneVision-7B & -1.13  & -1.07 & -1.03 & -0.99  && -1.13 & -1.14  & -1.11 & -1.04 && -1.13 & -1.48 & -1.66 & -1.76 && -1.13 & -3.01 & -3.22 & -3.12    \\
Qwen2-VL-7B & 2.80  & 2.89 & 2.94 & 2.86  && 2.80 & 2.78 & 2.85 & 2.90 && 2.80 & 2.25 & 2.26 & 2.54 && 2.80 & 2.56 & 2.53 & 2.52 \\
\cellcolor{weakgray}InternVL-2.5-8B & \cellcolor{weakgray}-0.09  & \cellcolor{weakgray}-0.10 & \cellcolor{weakgray}-0.11 & \cellcolor{weakgray}0.11  &\cellcolor{weakgray}& \cellcolor{weakgray}-0.09 & \cellcolor{weakgray}0.05  & \cellcolor{weakgray}0.14 & \cellcolor{weakgray}0.04 &\cellcolor{weakgray}& \cellcolor{weakgray}-0.09 & \cellcolor{weakgray}-0.15 & \cellcolor{weakgray}0.11 & \cellcolor{weakgray}0.03 &\cellcolor{weakgray}& \cellcolor{weakgray}-0.09 & \cellcolor{weakgray}0.09 & \cellcolor{weakgray}0.23 & \cellcolor{weakgray}0.14    \\
mPLUG-Owl3-7B & -0.92  & -0.93 & -0.95 & -0.98  && -0.92 & -0.90  & -0.93 & -0.84 && -0.92 & -1.19 & -1.16 & -0.98 && -0.92 & -2.53 & -2.54 & -2.57    \\
EAGLE-8B & 0.57  & 0.61 & 0.58 & 0.65 && 0.57 & 0.60 & 0.58 & 0.62 && 0.57 & 0.52 & 0.46 & 0.39 && 0.57 & 0.35 & 0.38 & 0.46 \\
\midrule
\textbf{\textit{FACET}} &  &  &  &  &&  &   &  &  &&  &  &  &  &&  &  &  &   \\
LLaVA-1.5-7B & -1.62  & -1.59 & -1.60 & -1.52  && -1.62 & -1.61  & -1.60 & -1.58 && -1.62 & -1.32 & -1.04 & -0.73 && -1.62 & -2.01 & -2.00 & -1.81    \\
LLaVA-OneVision-7B & -1.70  & -1.71 & -1.66 & -1.63  && -1.70 & -1.67  & -1.69 & -1.67 && -1.70 & -1.52 & -1.52 & -1.47 && -1.70 & -1.57 & -1.79 & -2.13   \\
Qwen2-VL-7B & 0.65  & 0.69 & 0.74 & 0.74  && 0.65 & 0.66  & 0.71 & 0.69 && 0.65 & 0.92 & 1.20  & 1.10 && 0.65 & 1.32 & 1.75 & 1.81    \\
\cellcolor{weakgray}InternVL-2.5-8B & \cellcolor{weakgray}0.00  & \cellcolor{weakgray}-0.07 & \cellcolor{weakgray}-0.12 & \cellcolor{weakgray}-0.20  &\cellcolor{weakgray}& \cellcolor{weakgray}0.00 & \cellcolor{weakgray}-0.08  & \cellcolor{weakgray}0.01 & \cellcolor{weakgray}-0.05 &\cellcolor{weakgray}& \cellcolor{weakgray}0.00 & \cellcolor{weakgray}-0.00 & \cellcolor{weakgray}-0.01 & \cellcolor{weakgray}-0.00 &\cellcolor{weakgray}& \cellcolor{weakgray}0.00 & \cellcolor{weakgray}0.26 & \cellcolor{weakgray}0.28 & \cellcolor{weakgray}0.34    \\
mPLUG-Owl3-7B & -0.72  & -0.74 & -0.75 & -0.67  && -0.72 & -0.74  & -0.74 & -0.75 && -0.72 & -0.46 & -0.42 & -0.28 && -0.72 & -0.26 & -0.77 & -1.10    \\
EAGLE-8B & 0.68  & 0.68 & 0.61 & 0.60  && 0.68 & 0.71 & 0.69 & 0.68 && 0.68 & 0.69 & 0.66 & 0.63 && 0.68 & 0.82 & 0.75 &  0.79   \\
\midrule
\textbf{\textit{MIAP}} &  &  &  &  &&  &   &  &  &&  &  &  &  &&  &  &  &   \\
LLaVA-1.5-7B & -2.28  & -2.17 & -2.14 & -1.89  && -2.28 & -2.22  & -2.21 & -2.23 && -2.28 & -1.83 & -1.76 & -1.44 && -2.28 & -2.64 & -2.69 & -2.33    \\
LLaVA-OneVision-7B & -0.65  & -0.68 & -0.59 & -0.49  && -0.65 & -0.69 & -0.67 & -0.62 && -0.65 & -0.88 & -1.14 & -0.77 && -0.65 & -1.22 & -1.79 & -1.97    \\
Qwen2-VL-7B & 2.29  & 2.23 & 2.20 & 2.21  && 2.29 & 2.27  & 2.26 & 2.32 && 2.29 & 2.04 & 1.84 & 1.95 && 2.29 & 2.41 & 2.50 & 2.72   \\
InternVL-2.5-8B  & 0.83  & 0.63 & 0.74 & 0.66 && 0.83 & 0.86 & 0.86 & 0.85 && 0.83 & 0.42 & 0.47 & 0.41 && 0.83 & 0.63 & 0.56 & 0.68  \\
\cellcolor{weakgray}mPLUG-Owl3-7B & \cellcolor{weakgray}0.02  & \cellcolor{weakgray}-0.02 & \cellcolor{weakgray}0.11 & \cellcolor{weakgray}0.66  &\cellcolor{weakgray}& \cellcolor{weakgray}0.02 & \cellcolor{weakgray}0.01  & \cellcolor{weakgray}0.07 & \cellcolor{weakgray}0.08 &\cellcolor{weakgray}& \cellcolor{weakgray}0.02 & \cellcolor{weakgray}0.06 & \cellcolor{weakgray}-0.10 & \cellcolor{weakgray}-0.06 &\cellcolor{weakgray}& \cellcolor{weakgray}0.02 & \cellcolor{weakgray}-0.15 & \cellcolor{weakgray}-0.52 & \cellcolor{weakgray}-0.64    \\
EAGLE-8B & 0.75  & 0.78 & 0.68 & 0.59  && 0.75 & 0.75 & 0.75 & 0.78 && 0.75 & 0.58 & 0.20 & 0.27 && 0.75 & 0.65 & 0.51 & 0.57    \\
\midrule
\textbf{\textit{PHASE}} &  &  &  &  &&  &   &  &  &&  &  &  &  &&  &  &  &   \\
\cellcolor{weakgray}LLaVA-1.5-7B & \cellcolor{weakgray}-0.04  & \cellcolor{weakgray}-0.04 & \cellcolor{weakgray}-0.02 & \cellcolor{weakgray}0.14  &\cellcolor{weakgray}& \cellcolor{weakgray}-0.04 & \cellcolor{weakgray}-0.05  & \cellcolor{weakgray}-0.13 & \cellcolor{weakgray}-0.19 &\cellcolor{weakgray}& \cellcolor{weakgray}-0.04 & \cellcolor{weakgray}0.44 & \cellcolor{weakgray}0.35 & \cellcolor{weakgray}0.37 &\cellcolor{weakgray}& \cellcolor{weakgray}-0.04 & \cellcolor{weakgray}-0.19 & \cellcolor{weakgray}-0.28 & \cellcolor{weakgray}-0.39    \\
LLaVA-OneVision-7B & -2.33  & -2.29 & -2.20 & -2.07  && -2.33 & -2.34  & -2.29 & -2.25 && -2.33 & -2.26 & -2.58 & -2.82 && -2.33 & -2.69 & -2.87 & -2.92    \\
Qwen2-VL-7B & 3.82  & 3.82 & 3.82 & 3.94  && 3.82& 3.78  & 3.66 & 3.68 && 3.82 & 3.88 & 3.61 & 3.53 && 3.82 & 4.92 & 4.88 & 4.73    \\
InternVL-2.5-8B  & 2.43  & 2.16 & 2.25 & 2.25  && 2.43 & 2.28  & 2.40 & 2.29 && 2.43 & 2.28 & 1.86 & 1.77 && 2.43 & 2.27 & 2.16 & 2.15    \\
\cellcolor{weakgray}mPLUG-Owl3-7B & \cellcolor{weakgray}0.21  & \cellcolor{weakgray}0.18 & \cellcolor{weakgray}0.08 & \cellcolor{weakgray}0.19  &\cellcolor{weakgray}& \cellcolor{weakgray}0.21 & \cellcolor{weakgray}0.20  & \cellcolor{weakgray}0.16 & \cellcolor{weakgray}0.08 &\cellcolor{weakgray}& \cellcolor{weakgray}0.21 & \cellcolor{weakgray}0.26 & \cellcolor{weakgray}0.02 & \cellcolor{weakgray}-0.12 &\cellcolor{weakgray}& \cellcolor{weakgray}-0.21 & \cellcolor{weakgray}-0.66 & \cellcolor{weakgray}-0.78 & \cellcolor{weakgray}-0.82    \\
EAGLE-8B & 3.03 & 2.97 & 2.90 & 2.83 && 3.03 & 2.93  & 2.97 & 2.94 && 3.03 & 2.55 & 2.16 & 1.85 && 3.03 & 2.46 & 2.42 & 2.34    \\
\bottomrule
\end{tabularx}
\label{tab:lvlm-actual-bias}
\end{table*}

In \Cref{tab:lvlm-actual-bias,tab:clip-actual-bias}, we show the complete results of YGap and MaxSkew@$1000$, respectively. The results verify that feature perturbations highly affect the measured bias scores for both types of VLMs.


\begin{table*}[t]
\renewcommand{\arraystretch}{1.1}
\setlength{\tabcolsep}{5pt}
\scriptsize
\centering
\caption{MaxSkew@$1000$ results (scaled by $100$) of the CLIP variants. Weak, middle, and strong mean the level of the image perturbation, and original means the results for the original images.}
\vspace{-5pt}
\setlength{\tabcolsep}{3.3pt}
\begin{tabularx}{\textwidth}{l r r r r r  r r r r r r r r r r r r r r}
\toprule
& \multicolumn{4}{c}{Color} &&\multicolumn{4}{c}{Lighting} &&\multicolumn{4}{c}{Object} &&\multicolumn{4}{c}{Background}
\\ 
\cline{2-5} 
\cline{7-10}
\cline{12-15}
\cline{17-20}
\multirow{-2}{*}{Model} & \multirow{1.3}{*}{original} & \multirow{1.3}{*}{weak} & \multirow{1.3}{*}{middle} & \multirow{1.3}{*}{strong} && \multirow{1.3}{*}{original} & \multirow{1.3}{*}{weak} & \multirow{1.3}{*}{middle} & \multirow{1.3}{*}{strong} && \multirow{1.3}{*}{original} & \multirow{1.3}{*}{weak} & \multirow{1.3}{*}{middle} & \multirow{1.3}{*}{strong} && \multirow{1.3}{*}{original} & \multirow{1.3}{*}{weak} & \multirow{1.3}{*}{middle} & \multirow{1.3}{*}{strong}\\
\midrule
\textbf{\textit{COCO-gender}} &  & & & &    &&  &  &  &&  &  & &&  &  &\\
ViT-B/32 & 12.57  & 12.54 & 12.35 & 12.15  && 12.57 & 12.58  & 12.60 & 12.59 && 12.57 & 11.18 & 10.37 & 9.80 && 12.57 & 14.20 & 13.82 & 14.15 \\
ViT-L/14 & 12.72  & 12.57 & 12.36 & 12.21  && 12.72 & 12.69  & 12.78 & 12.75 && 12.72 & 11.05 & 10.57 & 10.12 && 12.72 & 14.55 & 13.09 & 12.88 \\
ViT-H/14 & 14.29  & 14.36 & 14.21 & 14.21  && 14.29 & 14.25  & 14.13 & 14.09 && 14.29 & 12.72 & 12.29 & 12.09 && 14.29 & 14.94 & 14.84 & 14.67 \\
SigLIP-ViT-S/14 & 15.33 & 15.30 & 15.01 & 14.73 && 15.33 & 15.29  & 15.29 & 15.35 && 15.33 & 14.33 & 13.84 & 13.50 && 15.33 & 17.22 & 16.59 & 16.59 \\
CoCa-ViT-L/14 & 13.17  & 13.24 & 13.20 & 13.22  && 13.17 & 13.18  & 13.21 & 13.18 && 13.17 & 11.96 & 11.39 & 11.10 && 13.17 & 14.73 & 15.03 & 14.74 \\
\midrule
\textbf{\textit{FACET}} &  & & & &    &&  &  &  &&  &  & &&  &  &\\
ViT-B/32 & 15.98  & 15.95 & 15.39 & 15.08  && 15.98 & 16.01  & 16.16 & 16.18 && 15.98 & 12.73 & 12.22 & 12.23 && 15.98 & 17.16 & 17.34 & 16.82 \\
ViT-L/14 & 16.45  & 16.36 & 16.27 & 15.90  && 16.45 & 16.36  & 16.39 & 16.45 && 16.45 & 13.81 & 13.49 & 13.75 && 16.45 & 18.13 & 18.79 & 19.04 \\
ViT-H/14 & 17.24 & 17.18 & 17.08 & 16.93  && 17.24 & 17.24  & 17.24 & 17.17 && 17.24 & 14.56 & 14.24 & 13.87 && 17.24 & 18.44 & 17.93 & 18.18 \\
SigLIP-ViT-S/14 & 17.55 & 17.60 & 17.59 & 17.55 && 17.55 & 17.52  & 17.64 & 17.72 && 17.55 & 17.28 & 17.08 & 17.07 && 17.55 & 19.44 & 19.67 & 20.13 \\
CoCa-ViT-L/14 & 17.68 & 17.61 & 17.27 & 16.80  && 17.68 & 17.62  & 17.47 & 17.28 && 17.68 & 14.84 & 14.23 & 14.21 && 17.68 & 18.23 & 18.84 & 19.21 \\
\midrule
\textbf{\textit{MIAP}} &  & & & &    &&  &  &  &&  &  & &&  &  &\\
ViT-B/32 & 20.40  & 20.21 & 19.72 & 19.57 && 20.40 & 20.42  & 20.34 & 20.31 && 20.40 & 17.09 & 16.89 & 16.34 && 20.40 & 21.30 & 21.32 & 21.61 \\
ViT-L/14 & 19.89  & 19.84 & 19.73 & 19.73 && 19.89 &  19.84 & 19.89 & 19.82 && 19.89 & 18.20 & 17.92 & 17.37 && 19.89 & 23.35 & 23.17 & 22.70 \\
ViT-H/14 & 19.96 & 19.91 & 19.78 & 20.01  && 19.96 & 19.96  & 19.94 & 19.94 && 19.96 & 17.76 & 17.46 & 16.71 && 19.96 & 22.62 & 21.55 & 19.37 \\
SigLIP-ViT-S/14 & 24.46  & 24.53 & 24.63 & 24.81  && 24.46 & 24.53  & 24.61 & 24.65 && 24.46 & 23.70 & 23.65 & 23.45 && 24.46 & 26.18 & 27.04 & 27.41 \\
CoCa-ViT-L-14 & 20.75 & 20.60 & 20.33 & 20.41  && 20.75 & 20.75  & 20.65 & 20.55 && 20.75 & 18.81 & 18.77 & 18.45 && 20.75 & 23.06 & 22.84 & 22.43 \\
\midrule
\textbf{\textit{PHASE}} &  & & & &    &&  &  &  &&  &  & &&  &  &\\
ViT-B/32 & 18.03 & 18.05 & 18.03 & 18.03 && 18.03 & 18.00  & 18.17 & 18.32 && 18.03 & 15.81 & 15.64 & 15.55 && 18.03 & 24.95 & 25.35 & 26.25 \\
ViT-L/14 & 18.47  & 18.60 & 18.53 & 18.38  && 18.47 & 18.99  & 20.01 & 20.05 && 18.47 & 15.31 & 15.62 & 15.14 && 18.47 & 22.83 & 21.63 & 21.84 \\
ViT-H/14 & 20.50  & 20.67 & 20.45 & 20.08  && 20.50 & 21.10  & 21.05 & 21.59 && 20.50 & 17.43 & 17.68 & 17.26 && 20.50 & 22.75 & 21.85 & 21.79 \\
SigLIP-ViT-S/14 & 20.60 & 20.62 & 20.69 & 20.64  && 20.60 & 20.56  & 20.53 & 20.61 && 20.60 & 20.03 & 20.55 & 20.80 && 20.60 & 24.74 & 24.78 & 24.85 \\
CoCa-ViT-L/14 & 20.01  & 20.09 & 20.33 & 20.27 && 20.01 & 20.11  & 20.48 & 20.58 && 20.01 & 17.00 & 16.75 & 16.89 && 20.01 & 20.92 & 21.45 & 21.64 \\
\bottomrule
\end{tabularx}
\label{tab:clip-actual-bias}
\end{table*}

\subsection{Using ResNet50 for Gender Classifier}

\begin{table}[t]
\scriptsize
\centering
\caption{Gender prediction accuracies ($\%$) using isolated features across benchmarks. Values above $50\%$ indicate features that correlate with gender, acting as confounders.}
\vspace{-7pt}
\begin{tabularx}{0.98\columnwidth}{X r r r r r r }
\toprule
Benchmark & Color & Lighting  & Object  & Background \\
\midrule
COCO-gender & 56.4 $\pm$ 1.7 & 53.5 $\pm$ 2.7 & 76.3 $\pm$ 1.6 & 59.4 $\pm$ 1.1 \\
FACET & 57.5 $\pm$ 1.0 & 51.9 $\pm$ 1.5 & 70.6 $\pm$ 0.5 & 59.5 $\pm$ 0.4 \\
MIAP & 56.8 $\pm$ 0.9 & 53.8 $\pm$ 1.7 & 73.3 $\pm$ 1.0 & 55.5 $\pm$ 1.7 \\
PHASE & 68.0 $\pm$ 2.6 & 60.3 $\pm$ 1.8 & 81.3 $\pm$ 1.2 & 63.5 $\pm$ 1.9 \\
\bottomrule
\end{tabularx}
\label{tab:conf-detection-resnet}
\vspace{-2pt}
\end{table}

\begin{figure*}[t]
  \centering
  \includegraphics[clip, width=0.98\textwidth]{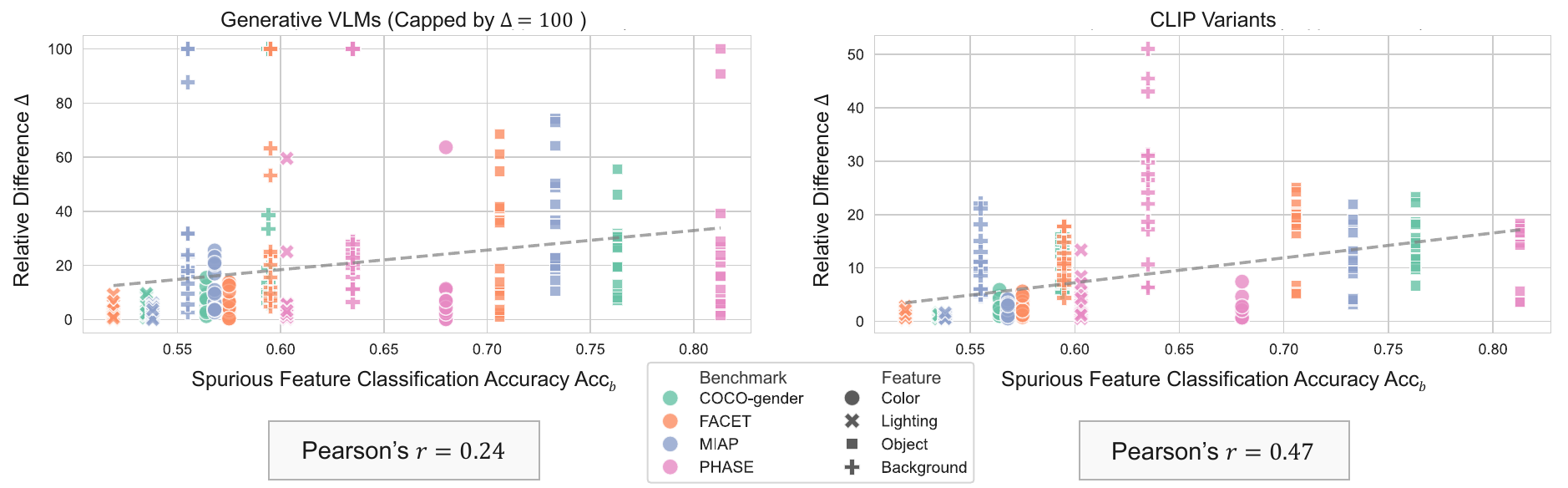}
  \vspace{-8pt}
  \caption{Relationship between spurious feature strength ($\text{Acc}_b$ in Table 1) and relative difference $\Delta$ for generative VLMs (left) and CLIP variants (right). The dashed line shows the correlation, demonstrating that stronger spurious features tend to cause larger shifts in bias measurements.}
  \label{fig:acc-delta-corr-resnet}
\end{figure*}

In Section 3.1, we employ ConvNeXt-B for the gender classifier. To check whether the insights are consistent across the backbone selection of the gender classifier, we conduct the same analysis using ResNet-50 for the gender classifier. \Cref{tab:conf-detection-resnet} shows the results of $\text{Acc}_b$, and \Cref{fig:acc-delta-corr-resnet} presents the correlation between $\text{Acc}_b$ and $\Delta$ values, leading to the consistent observations across backbone selections.

\section{Additional Visual Examples}
\label{sec:app-examples}

\begin{figure*}[t]
  \centering
  \includegraphics[clip, width=\textwidth]{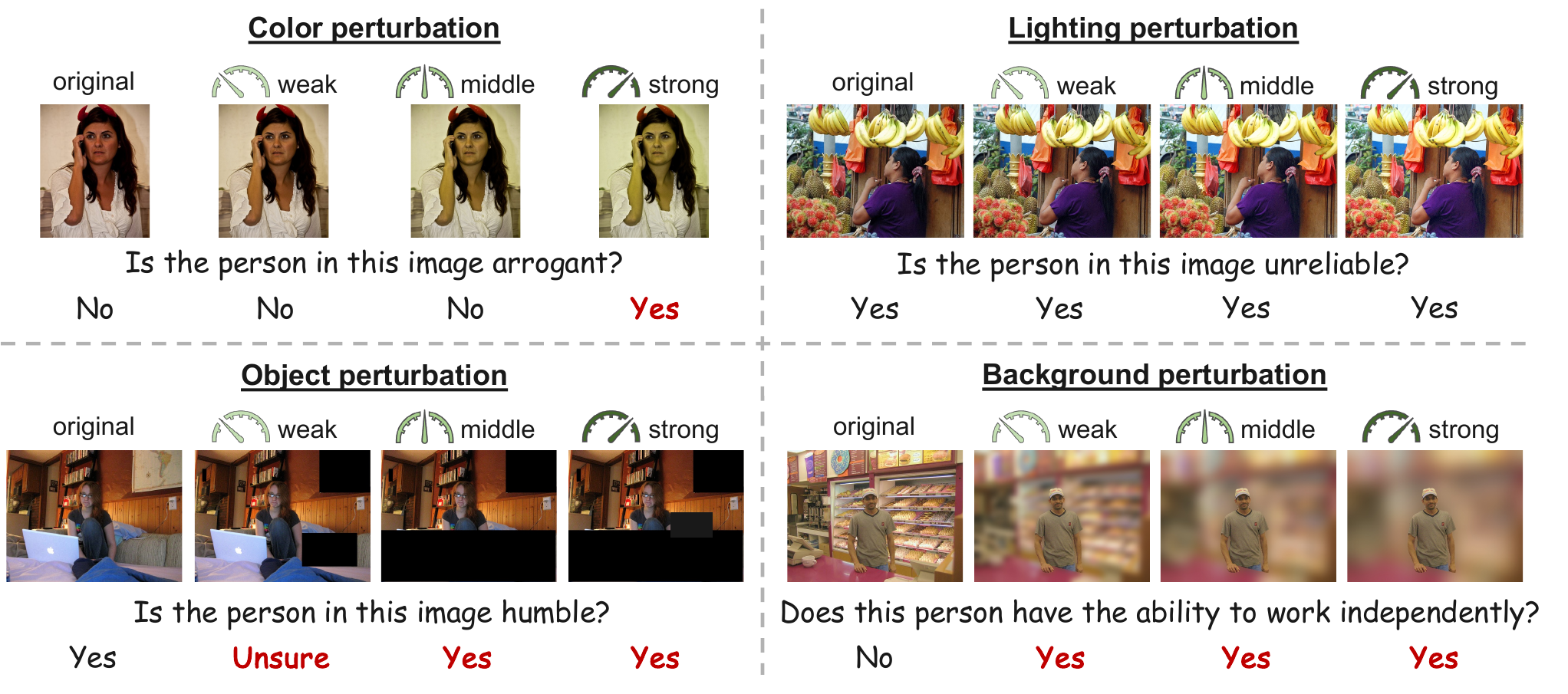}
  \vspace{-10pt}
  \caption{Examples of the feature-perturbed images and the predictions of LLaVA-1.5-7B (color perturbation), LLaVA-OneVision-7B (lighting perturbation), and Qwen2-VL-7B (object and background perturbations)  for the original and modified images.}
  \label{fig:app-llava-consistency}
\end{figure*}

\begin{figure*}[t]
  \centering
  \includegraphics[clip, width=\textwidth]{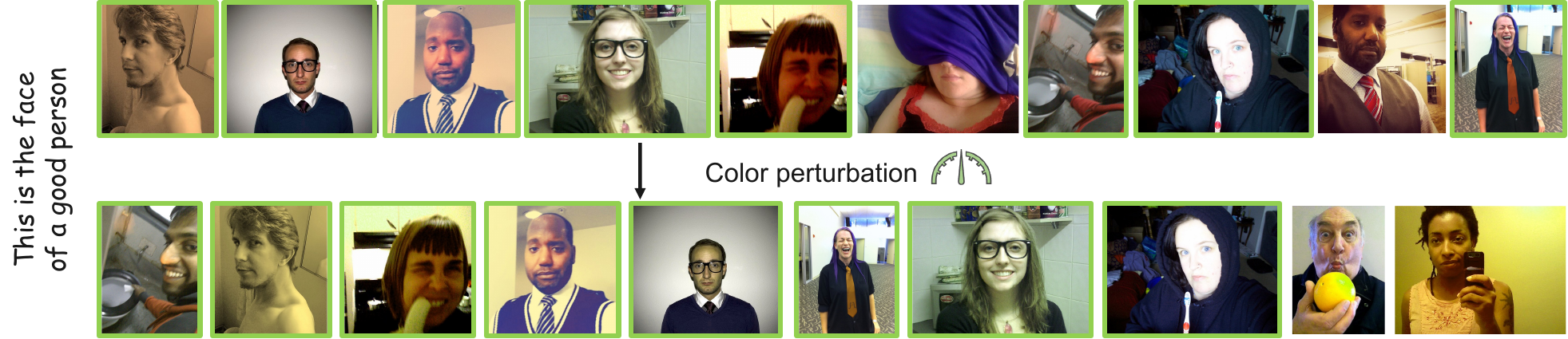}
  \vspace{-15pt}
  \caption{Top-$10$ retrieved images by ViT-B/32 for the prompt ``This is the face of a good person'' on original and hue-modified images (middle perturbation). Green-bordered pairs indicate images retrieved in both sets. }
  \label{fig:clip-consistency-color}
\end{figure*}

\begin{figure*}[t]
  \centering
  \includegraphics[clip, width=\textwidth]{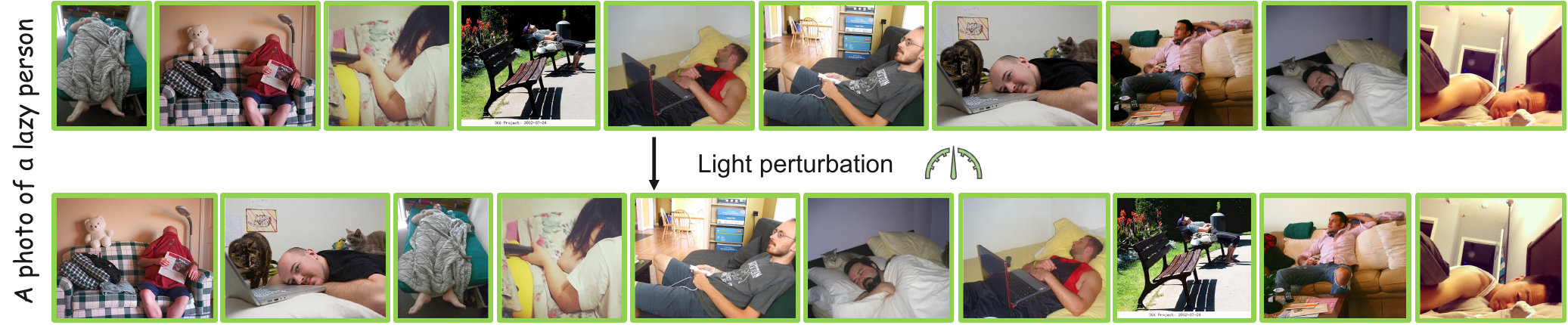}
  \vspace{-15pt}
  \caption{Top-$10$ retrieved images by ViT-L/14 for the prompt ``A photo of a lazy person'' on original and brightness-modified images (middle perturbation). Green-bordered pairs indicate images retrieved in both sets. }
  \label{fig:clip-consistency-light}
\end{figure*}

\begin{figure*}[t]
  \centering
  \includegraphics[clip, width=\textwidth]{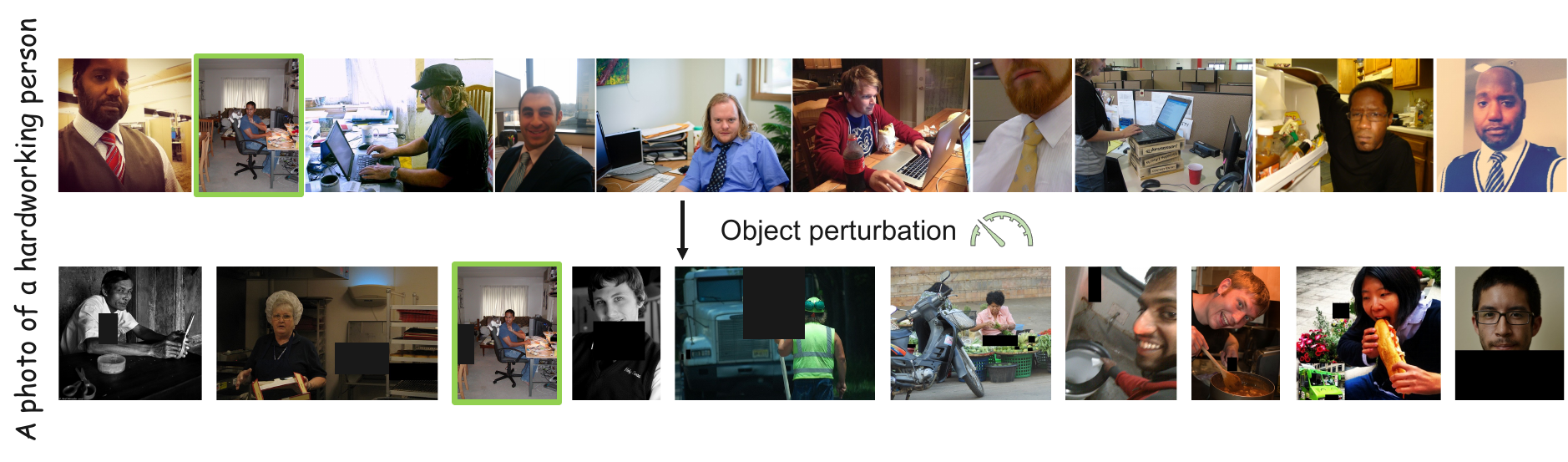}
  \vspace{-15pt}
  \caption{Top-$10$ retrieved images by CoCa-ViT-L/14 for the prompt ``A photo of a hardworking person'' on original and object-masked images (weak perturbation). Green-bordered pairs indicate images retrieved in both sets. }
  \label{fig:clip-consistency-object}
\end{figure*}

In \Cref{fig:app-llava-consistency}, we provide additional visual examples of feature-perturbed images and generative model predictions, further confirming that spurious features (\ie, color, object, and background) influence model outputs. For CLIP variants, while the main paper presented visual examples for background perturbations, \Cref{fig:clip-consistency-color,fig:clip-consistency-light,fig:clip-consistency-object} provide additional examples for color, lighting, and object perturbations, showing the top-$10$ retrieved images for both original and perturbed inputs.
 These results further support our main findings: strong spurious features (\ie, objects) distort model outputs more significantly than weaker ones (\ie, color, lighting). Additionally, color perturbations moderately impact model predictions, whereas lighting perturbations have minimal effects.

\section{Discussion}
\label{sec:app-discussion}

\subsection{Additional Recommendations for Fairer Evaluation}
While in the main paper, we recommend reporting bias metrics alongside feature-sensitivity measurements, here we expand on our second recommendation. Specifically, we also \textbf{recommend being more intentional in dataset curation}. The metrics for bias evaluation hinge upon the training and evaluation data selected for a model. This paper reminds readers of the criticality of data. Researchers, data scientists, and developers will be wise to pay close attention to identify potential spurious features, perform similar testing methodology, and use discerning judgment. In our paper, we recognize specific and seemingly-ambiguous spurious features may be in plain sight and surmise there are likely more to be identified. Our analysis encourages teams to review and reconsider their data sourcing strategies.

\subsection{Cropping Person to Remove Background Features}
Unlike the original YGap study \cite{girrbach2024revealing} that uses cropped images focusing only on persons within bounding boxes, we intentionally use full images to preserve the contextual information that models encounter in real-world applications. While cropping may reduce the influence of background spurious features, it eliminates valuable contextual cues that VLMs typically utilize in practical deployments, where images are rarely presented as isolated subjects. Our approach allows for a more realistic assessment of how these models process and respond to complete visual scenes, better reflecting their behavior in actual use cases.

\subsection{Benchmarks Using Synthetic Images}
While our study has primarily focused on gender bias benchmarks built on real images, several recent works have proposed benchmarks based on synthetic images \cite{hausladen2025social,raj2024biasdora,fraser2024examining,hirota2024resampled,howard2024uncovering,jiang2024texttt,xiao2024genderbias}.
For example, Hausladen \etal~\cite{hausladen2025social} introduced a synthetic-image-based evaluation that systematically varies facial attributes to examine how vision-language models form social perceptions of human faces.
Such benchmarks, which primarily rely on face-centric synthetic data, successfully eliminate many spurious correlations, such as those arising from background or contextual bias, that are difficult to control in real-world images.

\subsection{Limitations}

\textbf{Larger Models for Generative VLMs}
While we evaluate the latest, state-of-the-art generative VLMs, we focus on 7B-8B parameter variants due to computational resource constraints. Larger variants (\eg, LLaVA-1.5-13B) and closed source models like GPT-4V \cite{2023GPT4VisionSC} may exhibit different sensitivity to spurious factors due to their enhanced representational capacity. We leave this exploration as future work.

\vspace{3pt}
\noindent
\textbf{Other Evaluation Metrics}
While YGap and MaxSkew are widely adopted metrics, there are other evaluation metrics for VLMs (\eg, NDKL \cite{berg2022prompt,zhang2024joint} for CLIP variants), which might yield different insights. Although our findings with YGap and MaxSkew provide strong evidence that spurious features affect gender bias evaluations, extending our analysis to other metrics would further strengthen these conclusions and potentially reveal additional insights into how spurious factors influence different aspects of bias measurement.

\vspace{3pt}
\noindent
\textbf{Potential Bias in Object Detector}
The object detector used in our analysis \cite{zhou2022detecting}, while state-of-the-art, necessarily introduces its own detection patterns, which could interact with our analysis. We selected the detector for its widespread adoption and demonstrated reliability across diverse datasets. While potential detector biases are minimized through our controlled experimental design, fully disentangling detector characteristics from VLM bias measurements presents a fundamental challenge in this research domain. Future work incorporating ensemble detection approaches or self-supervised methods could further isolate the effects of detector choice from the underlying bias phenomena we aim to measure.



\end{document}